\documentclass[final]{cvpr}

\usepackage{times}
\usepackage{epsfig}
\usepackage{graphicx}
\usepackage{amsmath}
\usepackage{amssymb}
\usepackage[linesnumbered,ruled,vlined]{algorithm2e}
\usepackage{algpseudocode}
\usepackage{amsmath}
\usepackage{bm}
\usepackage{color}
\usepackage{makecell}
\usepackage{amsmath}
\usepackage{amsmath,subfigure,epsfig,bbding}
\usepackage[tableposition=top]{caption}
\usepackage[figuresright]{rotating}
\usepackage{threeparttable}
\usepackage{multirow}
\newcommand{\PreserveBackslash}[1]{\let\temp=\\#1\let\\=\temp}
\newcolumntype{C}[1]{>{\PreserveBackslash\centering}p{#1}}
\newcolumntype{R}[1]{>{\PreserveBackslash\raggedleft}p{#1}}
\newcolumntype{L}[1]{>{\PreserveBackslash\raggedright}p{#1}}

\usepackage[pagebackref=true,breaklinks=true,colorlinks,bookmarks=false]{hyperref}

\def\cvprPaperID{227} 
\def\confYear{CVPR 2021}

\begin{document}

\title{SDD-FIQA: Unsupervised Face Image Quality Assessment with \\Similarity Distribution Distance}
\author{Fu-Zhao Ou\textsuperscript{\rm \dag}\thanks{indicates equal contribution of this work. $\sharp$ denotes the corresponding author (e-mail: wangyg@gzhu.edu.cn). $\uplus$ indicates work done during an internship at Youtu Lab, Tencent. Code and model will be publicly available upon publication.} \  \textsuperscript{$\uplus$}, Xingyu Chen\textsuperscript{\rm \ddag $\ast$}, Ruixin Zhang\textsuperscript{\rm \ddag}, Yuge Huang\textsuperscript{\rm \ddag}, Shaoxin Li\textsuperscript{\rm \ddag}, \\ Jilin Li\textsuperscript{\rm \ddag}, Yong Li\textsuperscript{\rm \pounds}, Liujuan Cao\textsuperscript{\rm \S}, and Yuan-Gen Wang\textsuperscript{\rm \dag $\sharp$}\\ 
 $\qquad$  $\qquad$  \textsuperscript{\rm \dag}Guangzhou University $\qquad$ $\qquad$  \ \ \ \ $\qquad$ \textsuperscript{\rm \ddag}Youtu Lab, Tencent \\ \textsuperscript{\rm \pounds}Nanjing University of Science and Technology $\qquad$ \textsuperscript{\rm \S}Xiamen University\\
}

\maketitle

\begin{abstract}
In recent years, Face Image Quality Assessment (FIQA) has become an indispensable part of the face recognition system to guarantee the stability and reliability of recognition performance in an unconstrained scenario. 
For this purpose, the FIQA method should consider both the intrinsic property and the recognizability of the face image. 
Most previous works aim to estimate the sample-wise embedding uncertainty or pair-wise similarity as the quality score, which only considers the information from partial intra-class. However, these methods ignore the valuable information from the inter-class, which is for estimating to the recognizability of face image. In this work, we argue that a high-quality face image should be similar to its intra-class samples and dissimilar to its inter-class samples. Thus, we propose a novel unsupervised FIQA method that incorporates Similarity Distribution Distance for Face Image Quality Assessment (SDD-FIQA). Our method generates quality pseudo-labels by calculating the Wasserstein Distance (WD) between the intra-class similarity distributions and inter-class similarity distributions. With these quality pseudo-labels, we are capable of training a regression network for quality prediction. Extensive experiments on benchmark datasets demonstrate that the proposed SDD-FIQA surpasses the state-of-the-arts by an impressive margin. Meanwhile, our method shows good generalization across different recognition systems. 
\end{abstract}

\section{Introduction}
 Face recognition is one of the well-researched areas of biometrics \cite{He2005,LFW,ArcFace}. Under controlled conditions, the recognition system can usually achieve satisfactory performance. 
 However, in some real-world applications, recognition systems need to work under unconstrained environments (\emph{e.g.} surveillance camera and outdoor scenes), leading to significant degradation of recognition accuracy and unstable recognition performance. 
 Many researchers make progress in improving recognition accuracy under varying conditions  \cite{ArcFace,huang2020improving}, but sometimes the performance is still affected by the unpredictable environmental factors including pose, illumination, occlusion, and so on. 
 To keep the performance of face recognition system stable and reliable, Face Image Quality Assessment (FIQA) has been developed to support the recognition system to pick out high-quality images or drop low-quality ones for stable recognition performance.  \cite{gao2007standardization,aggarwal2011predicting}.\par

\begin{figure}
\centering
\setlength{\belowcaptionskip}{-0.18cm}
{\includegraphics[width=3.25in]{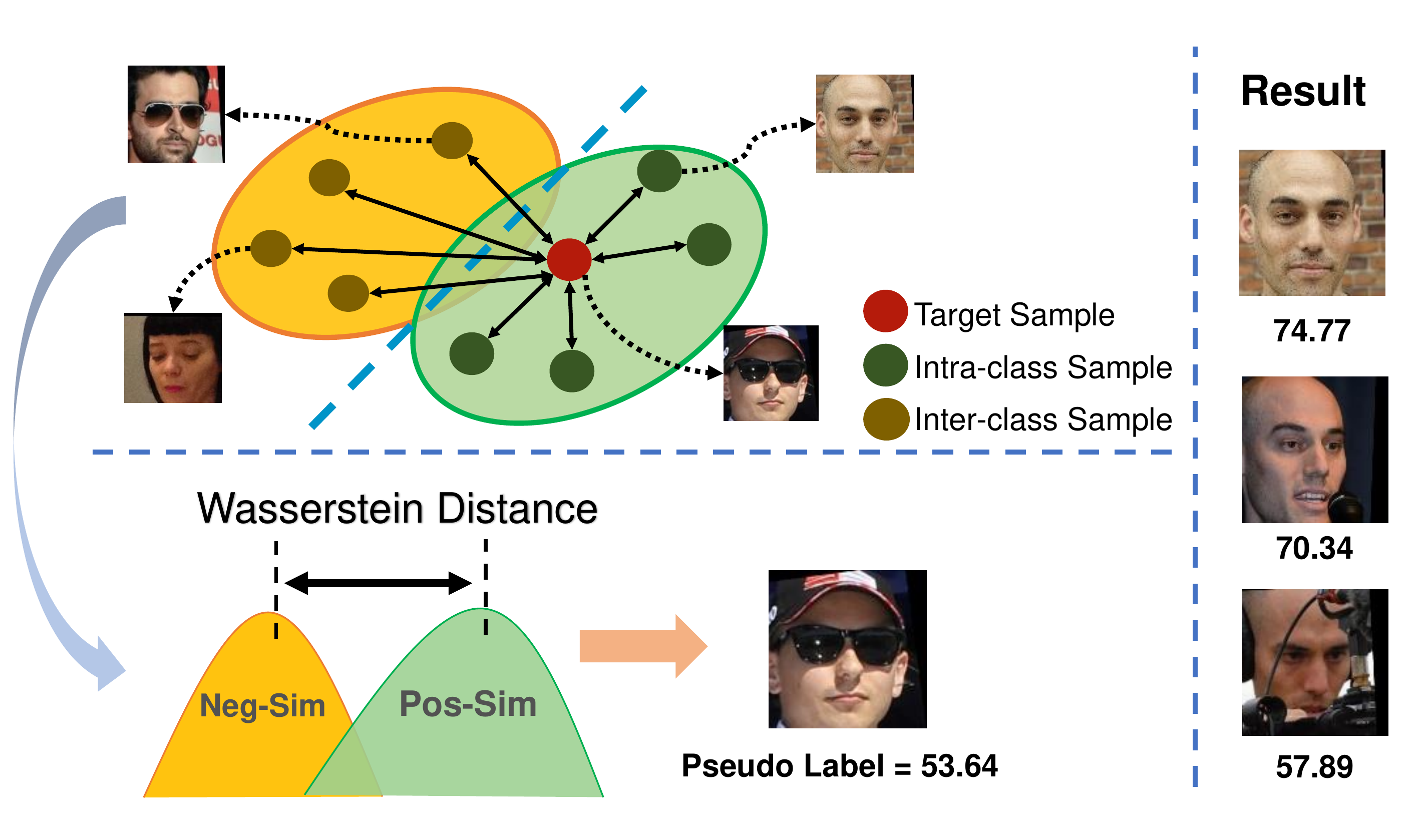}
\caption{Our method simultaneously considers similarities of the target sample (red point) with intra-class samples (green points), and with inter-class samples (yellow points). 
The distribution distance between Pos-Sim and Neg-Sim is calculated as the quality pseudo-label, which is more amicable to recognition performance. The visualized results are shown on the  right.}\label{SDD-FIQA}}
\end{figure}

\begin{figure*}
\centering
\setlength{\belowcaptionskip}{-0.18cm}
{\includegraphics[width=6.9in]{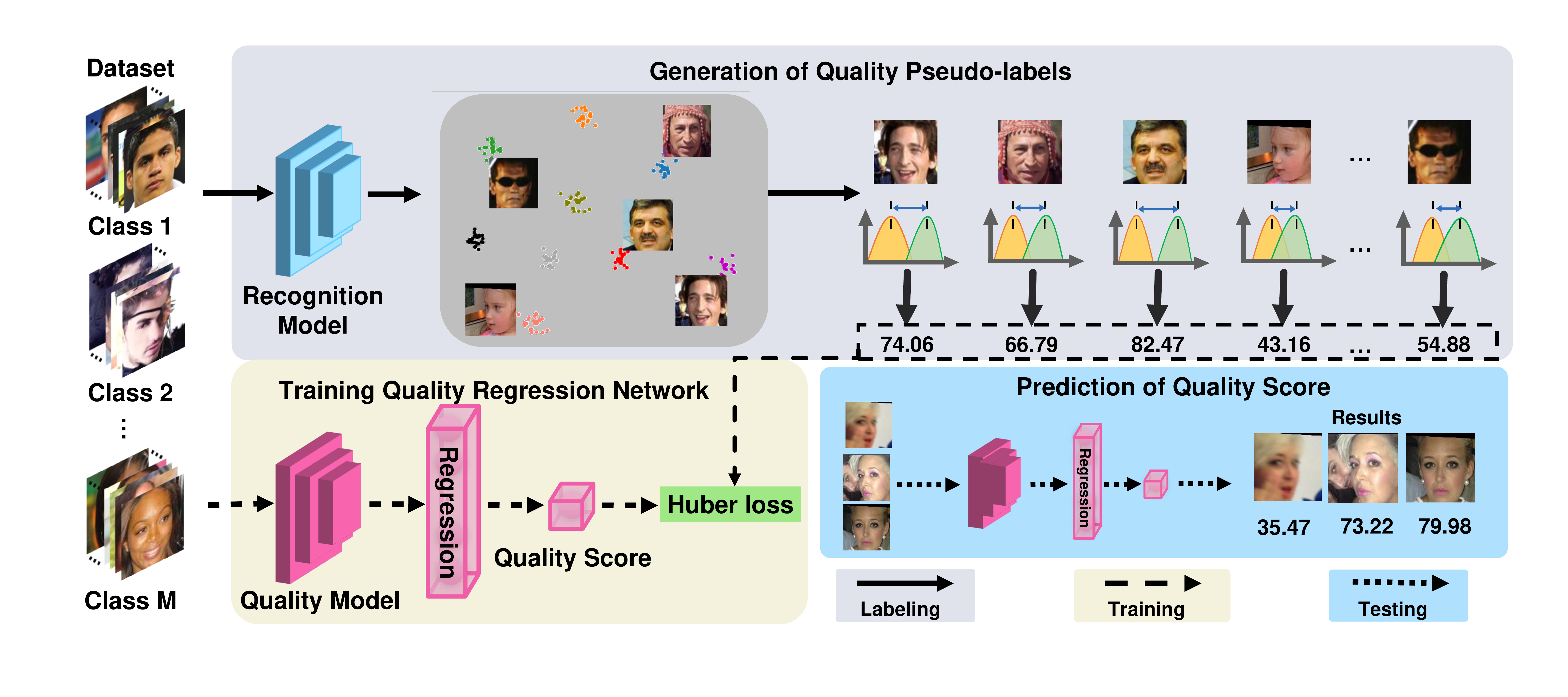}
\caption{The framework of SDD-FIQA. Step 1: The training data traverses the face recognition model, then its Pos-Sim and Neg-Sim are collected. Step 2: The Wasserstein distance between the Pos-Sim and Neg-Sim is calculated as the quality pseudo-label. Step 3: The quality regression network is trained under the constrain of Huber loss for FIQA. }
\label{Framework}}
\end{figure*}

Existing FIQA methods can be roughly categorized into two types: analytics-based  \cite{gao2007standardization,sellahewa2010image,wasnik2017assessing} and learning-based  \cite{aggarwal2011predicting,faceqnetv02019,PFE2019,SER-FIQ2020,PCNet2020}. 
Analytics-based FIQA defines quality metrics by Human Visual System (HVS) and evaluates face image quality with handcrafted features, such as asymmetries of facial area \cite{gao2007standardization}, illumination intensity \cite{sellahewa2010image}, and vertical edge density \cite{wasnik2017assessing}. However, these approaches have to manually extract features for different quality degradations, and it is unrealistic to annotate all possible degradations manually. 
Thus, more researchers take effort into the learning-based approach, which aims to generate quality scores directly from the recognition model without human effort. 
The most critical part of these approaches is to establish the mapping function between image quality and recognition model. 
Aggarwal \etal \cite{aggarwal2011predicting} proposed a multi-dimensional scaling approach to map space characterization features to genuine quality scores.  
Hernandez-Ortega \etal \cite{faceqnetv02019} calculate the Euclidean distance of intra-class recognition embeddings as the quality score.
Shi \etal \cite{PFE2019} and Terhorst \etal \cite{SER-FIQ2020} proposed to predict the variations of recognition embeddings as face image quality. 
Very recently, Xie \etal \cite{PCNet2020} proposed PCNet to evaluate face image quality via dynamically mining positive mated-pairs. 
Despite these learning-based methods make progress in FIQA, the performances are still unsatisfactory.  
Since they only consider the partial intra-class similarity or feature uncertainty from the recognition model and ignore the important information of inter-class similarity, which is the key factor for the recognizability of face image. 

In this paper, we propose a novel learning-based method called SDD-FIQA. Regarding FIQA as a recognizability estimation problem, we first reveal the intrinsic relationship between the recognition performance and face image quality. Specifically, for the target sample, we employ a recognition model to collect its intra-class similarity distributions (Pos-Sim) and inter-class similarity distributions (Neg-Sim). Then the Wasserstein Distance (WD) between these two distributions is calculated as the quality pseudo-label. Finally, a quality regression network is trained under the constraint of Huber loss. Our method can accurately predict the face image quality score in a label-free manner. The main idea of SDD-FIQA is shown in Fig. \ref{SDD-FIQA}, and the major contributions of this paper are as follows.
\begin{itemize}
\item We are the first to consider both the intra-class and inter-class recognition similarity distributions for the FIQA problem.
\item We propose a new framework of estimating face image quality, which is label-free and closely linked with the  recognition performance.
\item We are the first to evaluate the FIQA method by employing different recognition models for generalization evaluation, which is more consistent with the real-world applications. 
\item The proposed SDD-FIQA outperforms the state-of-the-arts by a large margin in terms of accuracy and generalization on benchmark datasets. 
\end{itemize}

\section{Related Work}
\subsection{Analytics-based FIQA}
Analytics-based FIQA approaches focus on analytically evaluating face image characteristics (such as pose, occlusion, and illumination) and define the quality metrics by HVS. For example, Gao \etal \cite{gao2007standardization} proposed a facial symmetry-based method. They measure the facial asymmetries caused by non-frontal lighting and improper facial pose. This method can quantize the impact of pose and illumination while failing to generalize to other quality factors. Sellahewa and Jassim \cite{sellahewa2010image} proposed an adaptive approach. They estimate the illumination quality by comparing the pixel-wise similarity between the reference image and the distorted one. However, such a method is sensitive to the background and can only deal with the illumination influence on image quality. Wasnik \etal \cite{wasnik2017assessing} evaluate the pose variation of face images based on vertical edge density and train a random forest model to predict the quality score. Their method requires a reference image, and the performance is not satisfactory under unconstrained conditions. Best-Rowden and Jain designed two FIQA methods with complete or partial human annotations \cite{Best2018}. However, human quality annotation is too labor-intensive and expensive to be applicable in practice.

\subsection{Learning-based FIQA}
Learning-based FIQA approaches aim to excavate the relationship between face image quality and recognition performance. Aggarwal \etal \cite{aggarwal2011predicting} were the first to propose the learning-based method. In their method, a space characterization features mapping model is constructed to predict the accuracy of recognition via multi-dimensional scaling. Wong \etal \cite{Wong2011} proposed an efficient patch-based method to obtain a similarity probabilistic model for FIQA. Chen \etal \cite{Chen2015} introduced a rank learning-based method. They use a ranking loss computed on multiple face images of an identity to train a network. 
Kim \etal \cite{Kim2015} take advantage of two factors, including visual quality and mismatch degree between training and testing images to predict face image quality. 
However, these methods can only work well with face images under controlled conditions. 

With the deep learning population, face representation using Deep Convolution Neural Network (DCNN) embedding has obtained impressive progress \cite{schroff2015facenet,wang2018cosface,huang2020improving}.  Researchers hope to directly estimate face image quality by DCNN embedding of face recognition. For example, Shi and Jain \cite{PFE2019} proposed the Probabilistic Face Embedding (PFE) method. Their method estimates the variance of element-wise embedding features and then applies the variance to FIQA. Hernandez-Ortega \etal \cite{faceqnetv02019} designed FaceQnet-V0. It calculates the Euclidean distance between the target face image and the best selection within intra-class as quality annotation. Furthermore, they developed FaceQnet-V1 \cite{faceqnetv12020} by adding three available face recognition systems to calculate Euclidean distance for quality annotation. Terhorst \etal \cite{SER-FIQ2020} proposed an unsupervised estimation of face image quality called SER-FIQ. They compute the mean Euclidean distance of the multiple embedding features from a recognition model with different dropout patterns as the quality score. Xie \etal \cite{PCNet2020} introduced the Predictive Confidence Network (PCNet). They apply pair-wise regression loss to train a DCNN for FIQA from intra-class similarity. Although these methods improve the versatility of FIQA by estimating the sample-wise uncertainty or computing pair-wise similarity, the relationship between intra-class and inter-class similarity has never been discovered, which is extremely important for the recognizability of face image.

\section{The Proposed SDD-FIQA}
The framework of the proposed SDD-FIQA is shown in Fig. \ref{Framework}. Assume $\mathcal{X}$, $\mathcal{Y}$,  and $\mathcal{F}$ denote the face image set, identity label set, and the recognition embedding set, respectively. We compose a triplet dataset $D = \{(x_1, y_1, f(x_1)), (x_2, y_2, f(x_2)), ..., (x_n, y_n, f(x_n))\} \subset \mathcal{X}$ $ \times \ \mathcal{Y} \times \mathcal{F}$. Let two samples $x_i, x_j$ from the same identity form a positive pair, two samples from different identities form a negative pair. For each training sample $x_i$, we denote $\mathcal{S}^P_{x_i}=\left\{s^P_{x_i}=\left\langle f(x_{i}), f(x_{j})\right\rangle \mid y_i=y_j\right\}$ and $\mathcal{S}^N_{x_i}=\left\{s^N_{x_i}=\left\langle f\left(x_{i}\right), f \left(x_{j}\right)\right\rangle \mid y_i\neq y_j \right\}$ as the similarity set of positive and negative pairs respectively, where $\langle f(x_i), f(x_j)\rangle$ denotes the cosine similarity between $f(x_i)$ and $f(x_j)$. 


\begin{figure}
\centering
\setlength{\belowcaptionskip}{-0.18cm}
{\includegraphics[width=3.2in]{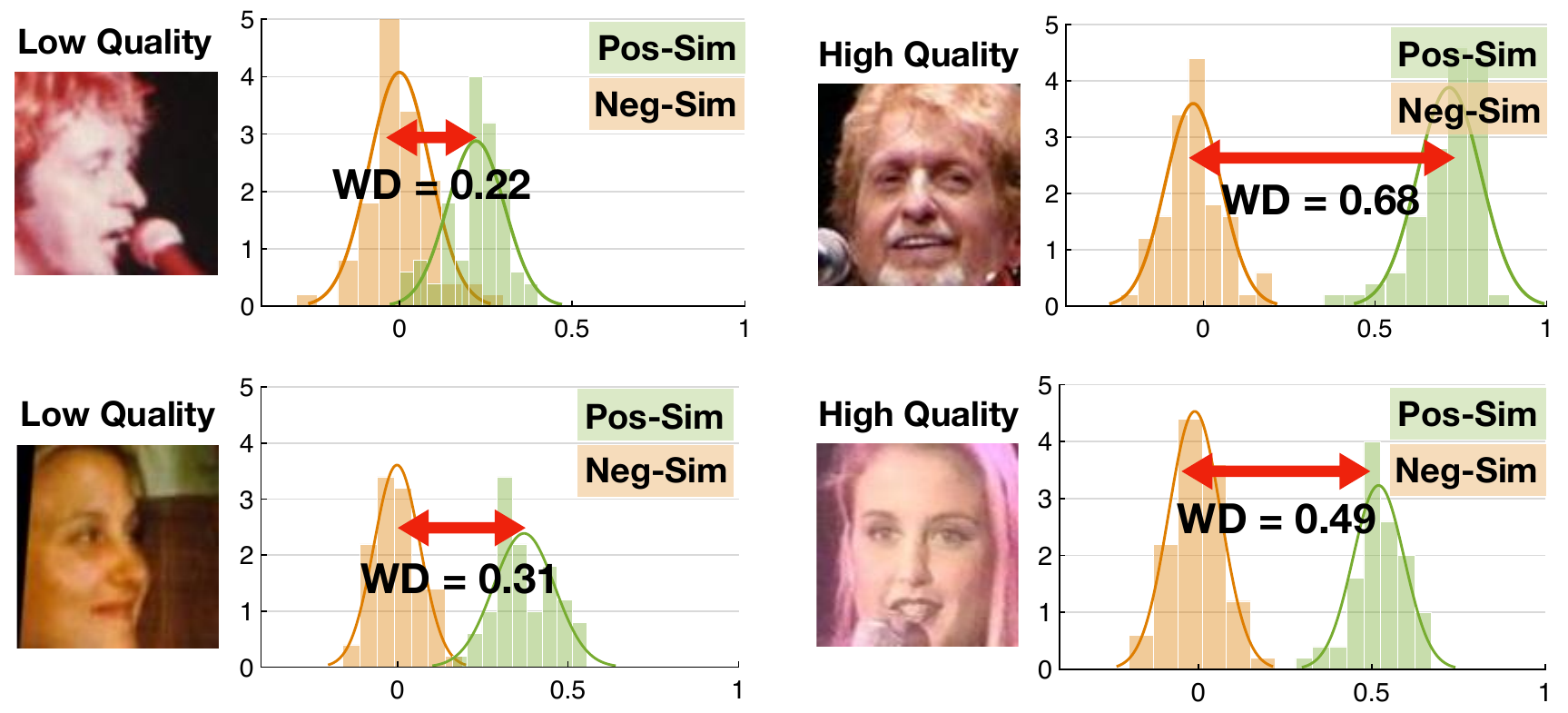}}
\caption{Illustration of the relationship between SDD and FIQ. The better FIQ, the larger distance between the Pos-Sim (colored in green) and Neg-Sim (colored in yellow).}
\label{Motivation}
\end{figure}


\subsection{Generation of Quality Pseudo-labels}
In this section, we describe the relationship between the face image quality and the recognition performance. We prove that such relationship can be deduced from the Error Versus Reject Curve (EVRC), which is a widely used metric in FIQA  \cite{Best2018,faceqnetv12020,SER-FIQ2020}. Specifically, assume that  $\mathcal{S}^P_{\mathcal{X}}=\{\mathcal{S}^P_{x_i}\}^n_{i=1}$ and $\mathcal{S}^N_{\mathcal{X}}=\{\mathcal{S}^N_{x_i}\}^n_{i=1}$ are the sets of all Pos-Sims and Neg-Sims, respectively. $\mathcal{S}^P_{\mathcal{X}|<\xi}$ denotes the subset of $\mathcal{S}^P_{\mathcal{X}}$ with similarity less than a threshold $\xi$.  $\mathcal{S}^N_{\mathcal{X}|>\xi}$ denotes the subset of $\mathcal{S}^N_{\mathcal{X}}$ with similarity greater than the threshold $\xi$, where $\xi \in [-1, 1]$. Then, False Match Rate (FMR) and False No-Match Rate (FNMR) can be respectively defined as
 \begin{equation}\label{FMR}
fmr = R_{fm}(\mathcal{X}, \xi)=\frac{|\mathcal{S}^N_{\mathcal{X}|>\xi}|}{|\mathcal{S}^N_{\mathcal{X}}|} 
 \end{equation}
 and
\begin{equation}\label{FNMR}
fnmr = R_{fnm}(\mathcal{X},  \xi)=\frac{|\mathcal{S}^P_{\mathcal{X}|<\xi}|}{|\mathcal{S}^P_{\mathcal{X}}|},
 \end{equation}
 where $|\cdot|$ denotes the set cardinality. Assume $\mathcal{X}^{Sub}_{|\sigma}$ is the subset of $\mathcal{X}$ with $\sigma$ percentage highest quality face samples. EVRC aims to measure the relationship between $\sigma$ and the FNMR of $\mathcal{X}^{Sub}_{|\sigma}$ at a fixed FMR. Note that since the FMR is required to be fixed, the threshold $\xi$ has to keep changing with respect to $\mathcal{X}^{Sub}_{|\sigma}$. According to the EVRC metric, we find that the image quality can be described by the gradient of the FNMR decrement. In other words, if a sample $x_i$ is dropped from the face dataset, the more sharply FNMR decreases, the lower quality of $x_i$ has. Inspired by this, we propose to compute the FNMR difference as the quality pseudo-label of $x_i$ (denoted as $Q_{x_{i}}$) by 
\begin{equation}\label{diff_fnmr}
Q_{x_{i}}=R_{fnm}(\mathcal{X}, \xi_{\mathcal{X}})-R_{fnm}(\mathcal{X}^{Sub}_{|-x_i}, \xi_{\mathcal{X}^{Sub}_{|-x_i}}),
\end{equation}
where $\mathcal{X}^{Sub}_{|-x_i}$ is the subset of $\mathcal{X}$ excluding $x_i$, $\xi_{\mathcal{X}}$ and $\xi_{\mathcal{X}^{Sub}_{|-x_i}}$ denote the thresholds for $\mathcal{X}$ and $\mathcal{X}^{Sub}_{|-x_i}$, respectively. We can see that for the given $\mathcal{X}$, $Q_{x_i}$ is only determined by $x_i$ and the threshold $\xi$. Since the FMR is fixed, the threshold $\xi$ can be calculated as $\xi=R_{fm}^{-1}(\mathcal{X}, fmr)$, where $R_{fm}^{-1}(\cdot)$ denotes the inverse function of $R_{fm}(\cdot)$. Objectively speaking, the actual quality of $x_i$ is independent of the FMR. However, by Eq. ~\eqref{diff_fnmr}, we can obtain only one empirical quality score under a fixed FMR. From a statistical point of view, we should take the expectation of $Q_{x_i}$ on the FMR to approach the actual quality score. Meanwhile, we can regard the FMR as a random variable  uniformly distributed over [0, 1]. Therefore, $Q_{x_i}$ can be reformulated as 
\begin{equation}\label{expect}
\begin{aligned}
Q_{x_i}=&\int_{0}^{1}  \lbrack R_{fnm}(\mathcal{X}, R_{fm}^{-1}(\mathcal{X}, fmr)) \\ 
& -R_{fnm}(\mathcal{X}^{Sub}_{|-x_i}, R_{fm}^{-1}(\mathcal{X}^{Sub}_{|-x_i}, fmr)) \rbrack d (fmr).
\end{aligned}
\end{equation}
By substituting Eq.~\eqref{FNMR} into Eq.~\eqref{expect}, we obtain
\begin{equation}\label{expfinal}
\begin{aligned}
Q_{x_{i}}=&\int_{0}^{1}\left[\frac{|\mathcal{S}^P_{\mathcal{X}|<R_{fm}^{-1}(\mathcal{X},fmr)}|}{|\mathcal{S}^P_{\mathcal{X}}|}\right]d(fmr) - \\
&\int_{0}^{1}\left[\frac{|\mathcal{S}^P_{\mathcal{X}^{Sub}_{|-x_i}|<R_{fm}^{-1}(\mathcal{X}^{Sub}_{|-x_i}, fmr)}|}{|\mathcal{S}^P_{\mathcal{X}^{Sub}_{|-x_i}}|}\right]d (fmr).
\end{aligned}
\end{equation}
According to our definition, it holds that $\mathcal{S}^P_{\mathcal{X}^{Sub}_{|-x_i}}=\mathcal{S}^P_{\mathcal{X}} - \mathcal{S}^P_{x_i}$ and $\mathcal{S}^N_{\mathcal{X}^{Sub}_{|-x_i}}=\mathcal{S}^N_{\mathcal{X}}-\mathcal{S}^N_{x_i}$. For a given dataset $\mathcal{X}$, both $\mathcal{S}^P_{\mathcal{X}}$ and $\mathcal{S}^N_{\mathcal{X}}$ are independent of the sample $x_i$. Therefore, the computation of Eq.~\eqref{expfinal} is uniquely determined by $\mathcal{S}^P_{x_i}$ and $\mathcal{S}^N_{x_i}$. Then Eq. ~\eqref{expfinal} can be simplified to
\begin{equation}\label{prove2}
 Q_{x_{i}} = \mathbf{F}(\mathcal{S}^P_{x_i}, \mathcal{S}^N_{x_i}),
 \end{equation}
where $\mathbf{F}(\cdot)$ is the mapping function from $\mathcal{S}^P_{x_i}$ and $\mathcal{S}^N_{x_i}$ to $Q_{x_{i}}$. In the following, we discover that the FIQ can be well described by SDD. Such discover can also be explained intuitively. For example, a high-quality face image is always easy to be recognized. That  means it is close to the intra-class samples and far from inter-class samples. In other words, the distance between the Pos-Sim and the Neg-Sim is large. Inversely, the low-quality face image produces a small SDD, as shown in \figref{Motivation}. Based on the above analysis, we propose to take advantage of the Wasserstein metric to measure SDD as $Q_{x_i}$, which is expressed by
\begin{equation}\label{wdistance}
\begin{aligned}
Q_{x_{i}} & = \mathbb{WD}(S_{x_{i}}^{P} \| S_{x_{i}}^{N}) \\
& = \inf _{\gamma \in \Pi(S_{x_{i}}^{P}, S_{x_{i}}^{N})} \mathbb{E}_{(s_{x_{i}}^{P}, s_{x_{i}}^{N}) \sim \gamma}\left[\left\|s_{x_{i}}^{P}-s_{x_{i}}^{N}\right\|\right],   \\
\end{aligned}
\end{equation}
where $\mathbb{WD}(\cdot)$ denotes Wasserstein distance, $\Pi(S_{x_{i}}^{P}, S_{x_{i}}^{N})$ denotes the set of all joint distribution $\gamma(s_{x_{i}},s_{y_{i}})$ whose marginals are respectively $S_{x_{i}}^{P}$ and $S_{x_{i}}^{N}$. 

According to the international biometric quality standard ISO/IEC 29794-1:2016 \cite{ISO2016}, the face quality pseudo-label should be regularized in the range [0,100] in which 100 and 0 indicate the highest and lowest quality pseudo-labels, respectively. Hence, the $Q_{x_i}$ is obtained by  
\begin{equation}\label{eq3.4}
Q_{x_i}=\delta[\mathbb{WD}(S_{x_{i}}^{P} \| S_{x_{i}}^{N})] ,
\end{equation}
where
\begin{equation}\label{eq3.5}
\delta[l_{x_i}]=100 \times \frac{l_{x_i} - min(\mathcal{L})}{max(\mathcal{L})-min(\mathcal{L})}, l_{x_i}\in\mathcal{L} ,
\end{equation}
where $l_{x_i}$ = $\mathbb{WD}(S_{x_{i}}^{P} \| S_{x_{i}}^{N})$, and $\mathcal{L} = \{l_{x_i} | i=1, 2, ..., n\}$.

\subsection{Acceleration of Label Generation}\label{3.2}
For a face dataset of size $n$, if the pair-wise similarity is taken into consideration overall dataset, the time complexity of $Q_{x_i}$ is $\mathcal{O}(n^2)$. This is high computation when encountering a large-scale dataset. To reduce the computational complexity, we randomly select $m$ positive pairs and $m$ negative pairs, where $m$ is set to be even and $m \ll n$. Then $Q_{x_i}$ is computed $K$ times, and Eq. ~\eqref{eq3.4} can be transformed to 
\begin{equation}\label{eq3.6}
\tilde{Q}_{x_i}=\frac{\sum_{k=1}^K\delta[\mathbb{WD}(S_{x_{i}^k}^{P_m} \| S_{x_{i}^k}^{N_m})]}{K},
\end{equation}
where $S_{x_{i}^k}^{P_m}$ and $S_{x_{i}^k}^{N_m}$ are the sampled similarities of positive and negative pairs within $m$ pairs, respectively. Note that for each sample, since $m$ and $K$ are set to be constant, the computational complexity of Eq.~\eqref{eq3.6} is $\mathcal{O}(2m \times K)$, where $2m \times K$ is at least one order lower than $n$. Therefore, the generation of quality annotation over all dataset achieves $\mathcal{O}(2m \times K \times n) = \mathcal{O}(n)$ time complexity. Suppose that $\delta[\mathbb{WD}(S_{x_{i}^k}^{P} \| S_{x_{i}^k}^{N})]$ is a noiseless estimation of $Q_{x_i}$ and a noise variable $\epsilon_k$ is induced in each sampling. Then Eq. ~\eqref{eq3.6} can be rewritten as 
\begin{equation}\label{eq3.7}
\tilde{Q}_{x_i}=\delta[\mathbb{WD}(S_{x_{i}}^{P} \| S_{x_{i}}^{N})] = \frac{\sum_{k=1}^K(\delta[\mathbb{WD}(S_{x_{i}}^{P_m} \| S_{x_{i}}^{N_m})]+\epsilon_k)}{K}.
\end{equation}
Note that $\epsilon_k$ can be regarded as the difference between $\delta[\mathbb{WD}(S_{x_{i}}^{P} \| S_{x_{i}}^{N})]$ and $\delta[\mathbb{WD}(S_{x_{i}^k}^{P_m} \| S_{x_{i}^k}^{N_m})]$. According to the probability theory \cite{Papoulis2002}, we have $\lim\limits_{k \rightarrow \infty} \sum_{i=1}^{K} \epsilon_{i}=0$. Hence, Eq. ~\eqref{eq3.6} is an unbiased estimation of Eq. ~\eqref{eq3.4}.



\subsection{Quality Regression Network}

With our generated quality pseudo-labels, we can train an individual FIQA model, which no longer resorts to the recognition system to output quality score. To match the predictions of the quality regression network with the recognition system, knowledge-transfer is applied during training. Specifically, we first remove the embedding and classification layers from a pre-trained face recognition model. Then we employ a dropout operator with 0.5 probability to avoid overfitting during training, and add a fully connected layer in order to output the quality score for FIQA. Finally, we use the Huber loss function \cite{Huber} to train the quality regression network, which is given by
\begin{equation}\label{huber}
\begin{aligned}
L_{\zeta}(&x_i, \tilde{Q}_{x_i}; \pi) = \\
& \begin{cases}
\frac{1}{2}\left\|\tilde{Q}_{x_i}-\phi_{\pi}\left(x_{i}\right)\right\|_{2}, &\mbox{if }\left|y_{i}-\phi_{\pi}\left(x_{i}\right)\right| \leq \zeta\\
\zeta\left\|\tilde{Q}_{x_i}-\phi_{\pi}\left(x_{i}\right)\right\|-\frac{1}{2} \zeta^{2},  &\mbox{otherwise},
\end{cases}
\end{aligned}
\end{equation}
where $\zeta$ is the location parameter, $\pi$ and $\phi_{\pi}(\cdot)$ denote the network parameters and the regression output, respectively. \par

Compared with existing FIQA methods, our method considers both Pos-Sim and Neg-Sim overall dataset, which is more amicable to recognition performance. Moreover, our approach is unsupervised without any human annotation.

\section{Experiments}
\subsection{Experimental Setup}
\noindent\textbf{Datasets}. The refined MS-Celeb-1M (MS1M) \cite{ArcFace} is utilized as the training data of the recognition model and quality regression model. Besides, CASIA-WebFace (CASIA) \cite{CASIA} is adopted to train the recognition model for  evaluating the generalization of our method. During testing, LFW \cite{LFW}, Adience \cite{Adience}, UTKFace \cite{UTKFace}, and IJB-C \cite{IJBC} are employed as test datasets. Note that IJB-C is one of the most challenging public benchmarks and contains large variations in pose, expression, illumination, blurriness, and occlusion. 

\begin{table}
\setlength{\abovecaptionskip}{0.05cm}
\setlength{\belowcaptionskip}{-0.1cm}
\begin{center}
\caption{AOC results on same recognition model setting.}\label{ACC_Res50_MS1M}
\resizebox{236pt}{135pt}{
\begin{tabular}{|C{0.7cm}|c|c|c|c|c|}
\Xhline{0.6pt}
$ $	 &\bf{LFW}	&\bf{FMR}=$\bm{1e^{-2}}$ 	&\bf{FMR=}$\bm{1e^{-3}}$		&\bf{FMR=}$\bm{1e^{-4}}$ 	&\bf{Avg}\\
\Xhline{0.6pt}
\multirow{3}*{\rotatebox{90}{\makecell{\textbf{\small{Analytics}}\\ \textbf{\small{Based}}}}}
& BRISQUE \cite{BRISQUE}				&0.0321   &0.0953   &0.2579   &0.1284 \\
& BLIINDS-II \cite{BLIINDS-II}		&0.1755   &0.0923   &0.2270   &0.1649 \\
& PQR \cite{PQR}						&0.1921   &0.2952   &0.3519   &0.2797 \\
\hline
\multirow{5}*{\rotatebox{90}{\makecell{\textbf{\small{Learning}}\\ \textbf{\small{Based}}}}}
& FaceQnet-V0 \cite{faceqnetv02019} 	&0.4535   &0.4955   &0.5399   &0.4963 \\ 
& FaceQnet-V1 \cite{faceqnetv12020} 	&0.4417   &0.5471   &0.6167   &0.5352 \\
& PFE \cite{PFE2019}					&0.4814   &0.5057   &0.5895   &0.5255 \\ 
& SER-FIQ \cite{SER-FIQ2020} 		&0.5669   &0.6675   &0.7469   &0.6604 \\ 
& PCNet \cite{PCNet2020} 		&0.6975   &0.7275   &0.7197   &0.7149 \\ 
\Xhline{0.6pt}
& \bf{SDD-FIQA (Our)} 						&\bf{0.8284}   &\bf{0.7993}   &\bf{0.8170}   &\bf{0.8149} \\ 
\hline
\hline
\hline
$ $	 &\bf{Adience}	&\bf{FMR=}$\bm{1e^{-2}}$ 	&\bf{FMR=}$\bm{1e^{-3}}$		&\bf{FMR=}$\bm{1e^{-4}}$ 	&\bf{Avg}\\
\Xhline{0.6pt}
\multirow{3}*{\rotatebox{90}{\makecell{\textbf{\small{Analytics}}\\ \textbf{\small{Based}}}}}
& BRISQUE \cite{BRISQUE}				&0.2686   &0.2056   &0.2353   &0.2365 \\
& BLIINDS-II \cite{BLIINDS-II}		&0.1723   &0.1634   &0.1565   &0.1640 \\
& PQR \cite{PQR}						&0.2454   &0.2102   &0.1962   &0.2173 \\
\hline
\multirow{5}*{\rotatebox{90}{\makecell{\textbf{\small{Learning}}\\ \textbf{\small{Based}}}}}
& FaceQnet-V0 \cite{faceqnetv02019} 	&0.4756   &0.5021   &0.4735   &0.4837 \\ 
& FaceQnet-V1 \cite{faceqnetv12020} 	&0.3809   &0.4613   &0.4350   &0.4257 \\
& PFE \cite{PFE2019}					&0.5490   &0.6046   &0.5556   &0.5698 \\
& SER-FIQ \cite{SER-FIQ2020}  		&0.5009   &0.5539   &0.4384   &0.4977 \\ 
& PCNet \cite{PCNet2020} 		&0.5224   &0.5597   &0.5255   &0.5359 \\ 
\Xhline{0.6pt}
& \bf{SDD-FIQA (Our)} 						&\bf{0.5962}   &\bf{0.6307}   &\bf{0.5719}   &\bf{0.5996} \\
\hline
\hline
\hline
$ $	 &\bf{IJB-C}	&\bf{FMR=}$\bm{1e^{-2}}$ 	&\bf{FMR=}$\bm{1e^{-3}}$		&\bf{FMR=}$\bm{1e^{-4}}$ 	&\bf{Avg}\\
\Xhline{0.6pt}
\multirow{3}*{\rotatebox{90}{\makecell{\textbf{\small{Analytics}}\\ \textbf{\small{Based}}}}}
& BRISQUE \cite{BRISQUE}				&0.2943   &0.3292   &0.4216   &0.3484 \\
& BLIINDS-II \cite{BLIINDS-II}		&0.3641   &0.3656  &0.3806  &0.3701\\
& PQR \cite{PQR}						&0.4991   &0.4979   &0.5230   &0.5067 \\
\hline
\multirow{5}*{\rotatebox{90}{\makecell{\textbf{\small{Learning}}\\ \textbf{\small{Based}}}}}
& FaceQnet-V0 \cite{faceqnetv02019} 	&0.6149   &0.6047   &0.6330   &0.6175 \\ 
& FaceQnet-V1 \cite{faceqnetv12020} 	&0.6343  &0.6332   &0.6552   &0.6409 \\
& PFE \cite{PFE2019}					&0.6899   &0.6964   &0.7306   &0.7056 \\
& SER-FIQ \cite{SER-FIQ2020}  		&0.6115   &0.5976   &0.6135   &0.6075 \\ 
& PCNet \cite{PCNet2020} 		&0.7053   &0.7055   &0.7363   &0.7157 \\ 
\Xhline{0.6pt}
& \bf{SDD-FIQA (Our)} 						&\bf{0.7209}   &\bf{0.7239}   &\bf{0.7539}   &\bf{0.7329} \\ 
\Xhline{0.6pt}
\end{tabular}}
\end{center}
\end{table}

 \begin{figure*}[htbp]
 \centering
 \subfigure[]{
 \includegraphics[width=5.75cm]{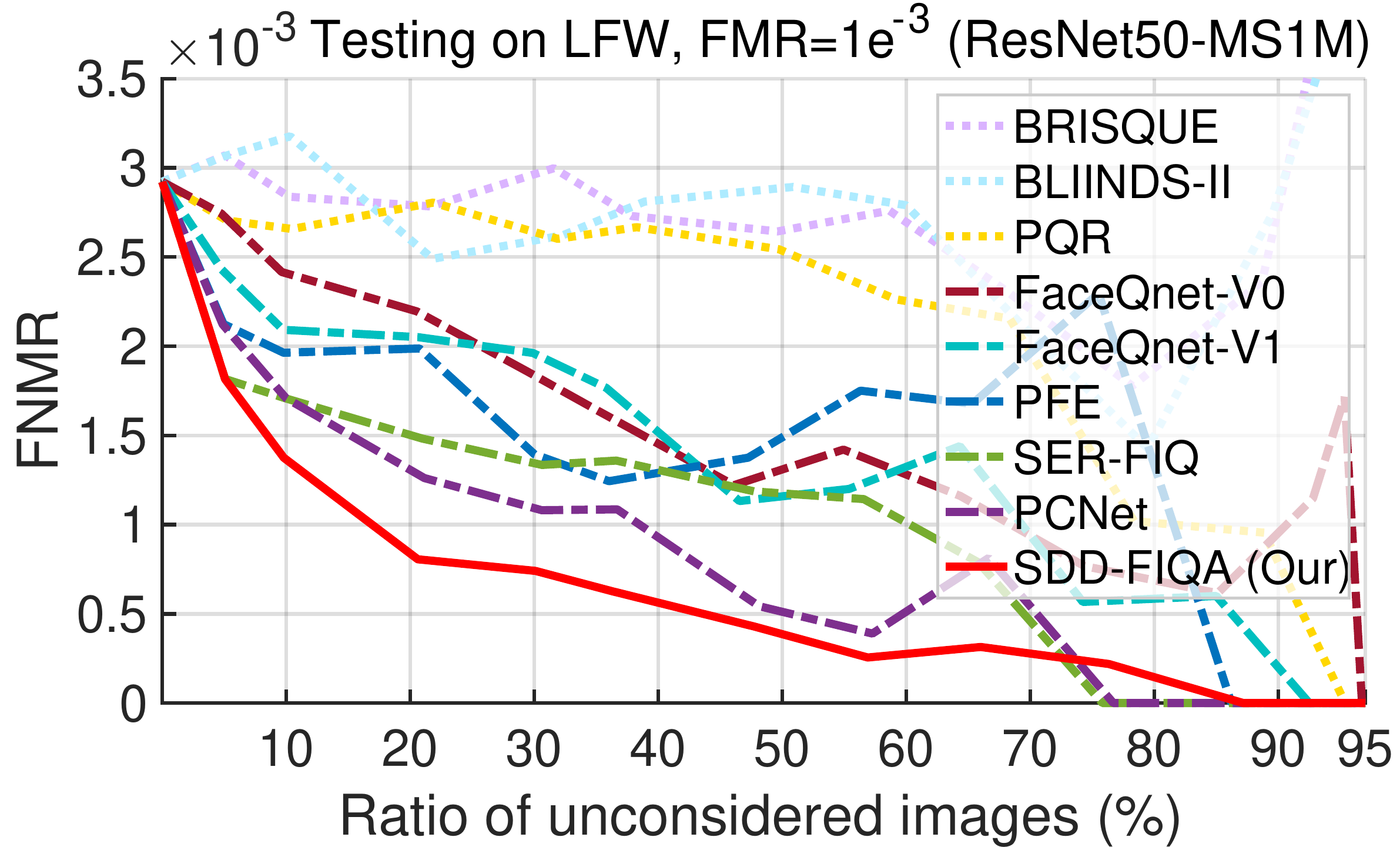}
 }\hspace{-6mm}
 \quad
 \subfigure[]{
 \includegraphics[width=5.75cm]{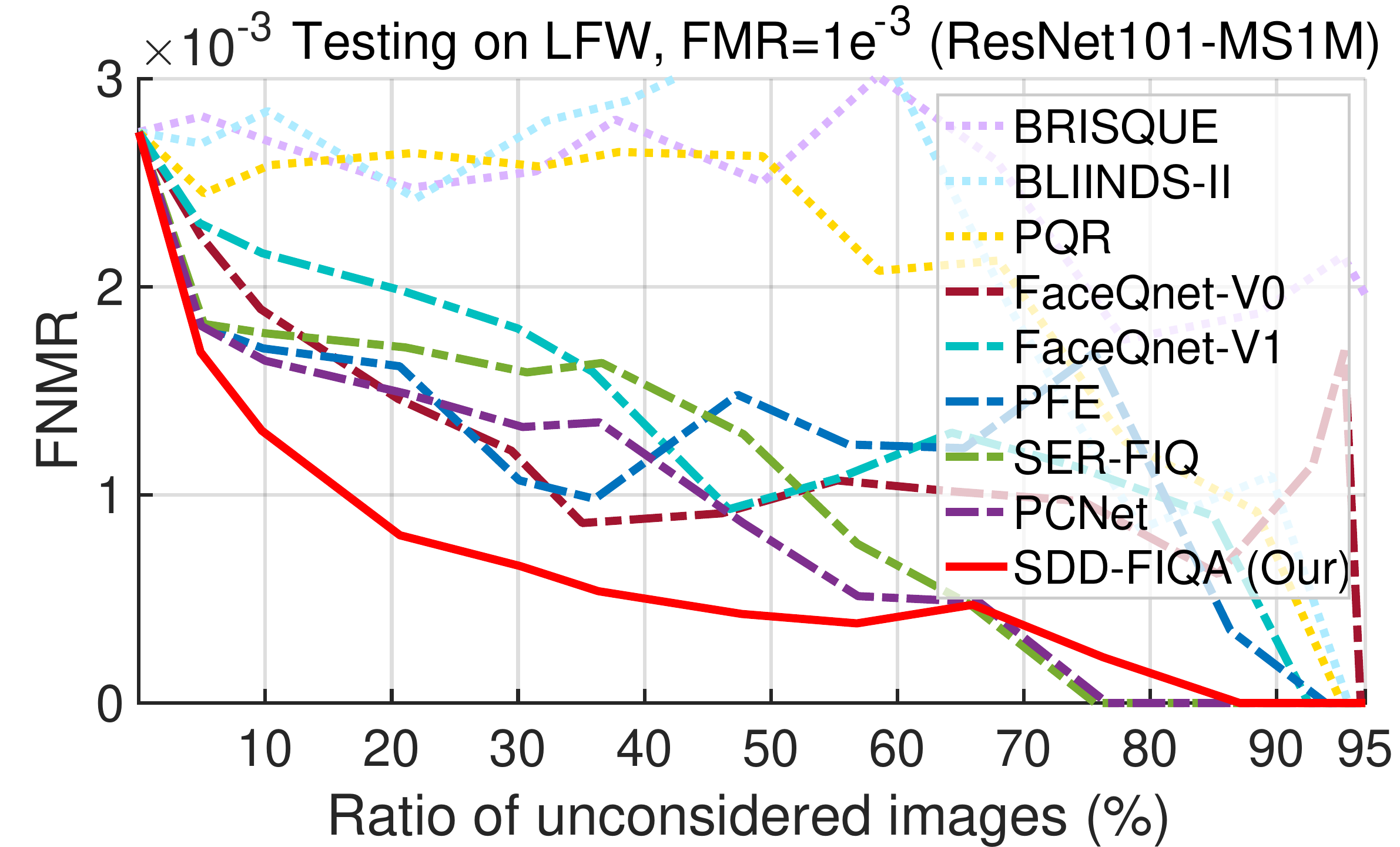}
 }\hspace{-6mm}
 \quad
 \subfigure[]{
 \includegraphics[width=5.75cm]{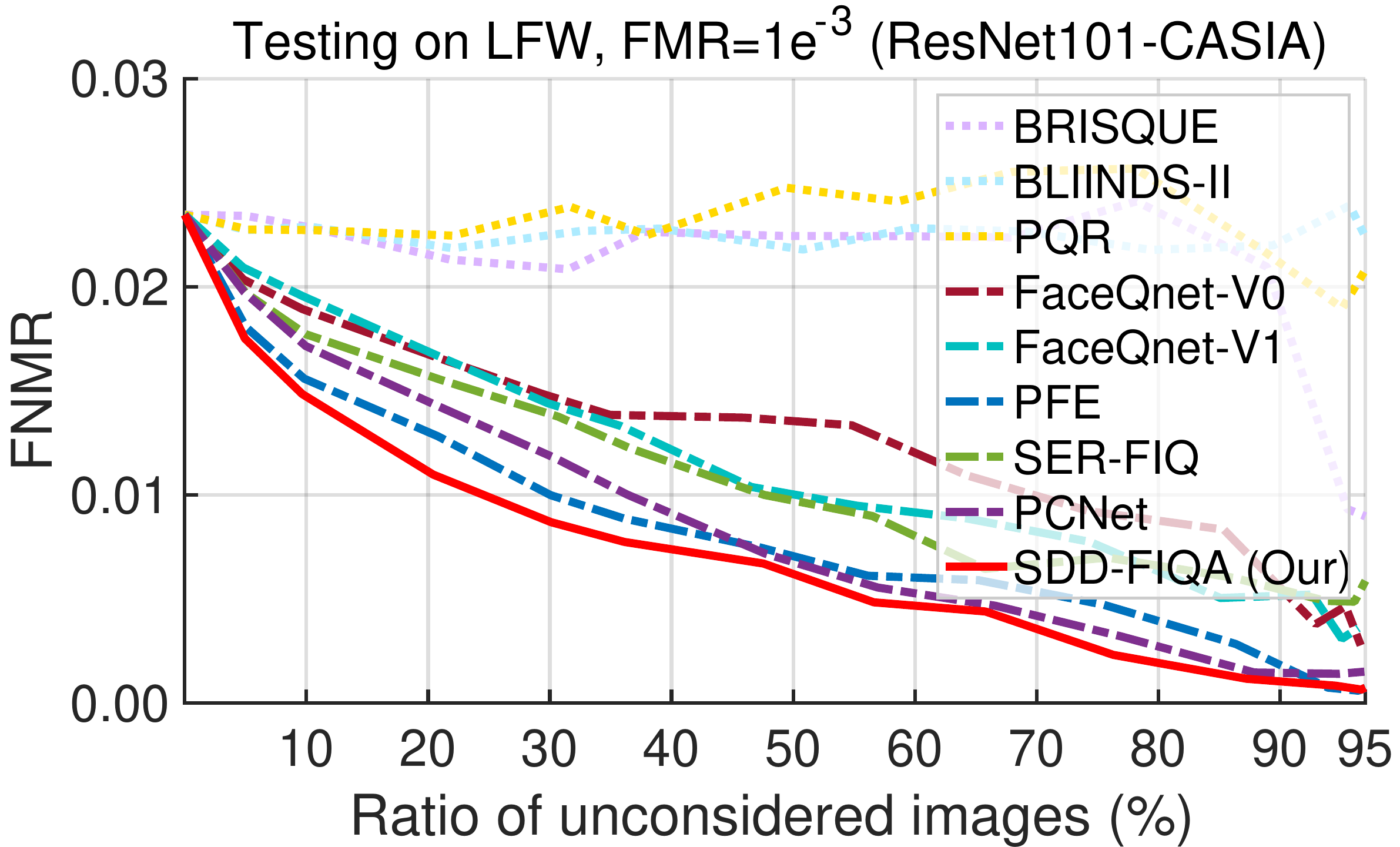}
 }\hspace{-6mm}
 \quad
 \subfigure[]{
 \includegraphics[width=5.75cm]{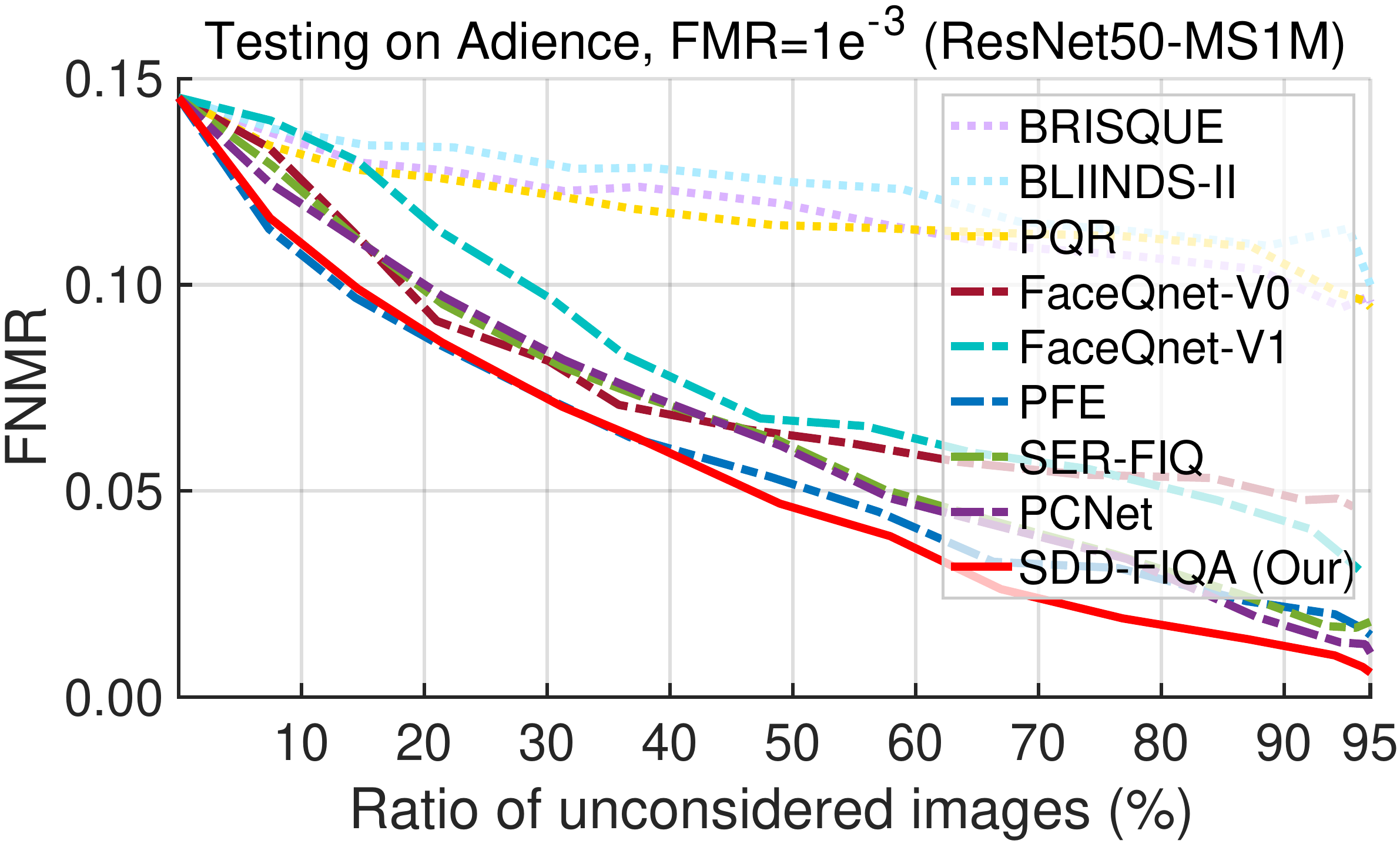}
 }\hspace{-6mm}
 \quad
 \subfigure[]{
 \includegraphics[width=5.75cm]{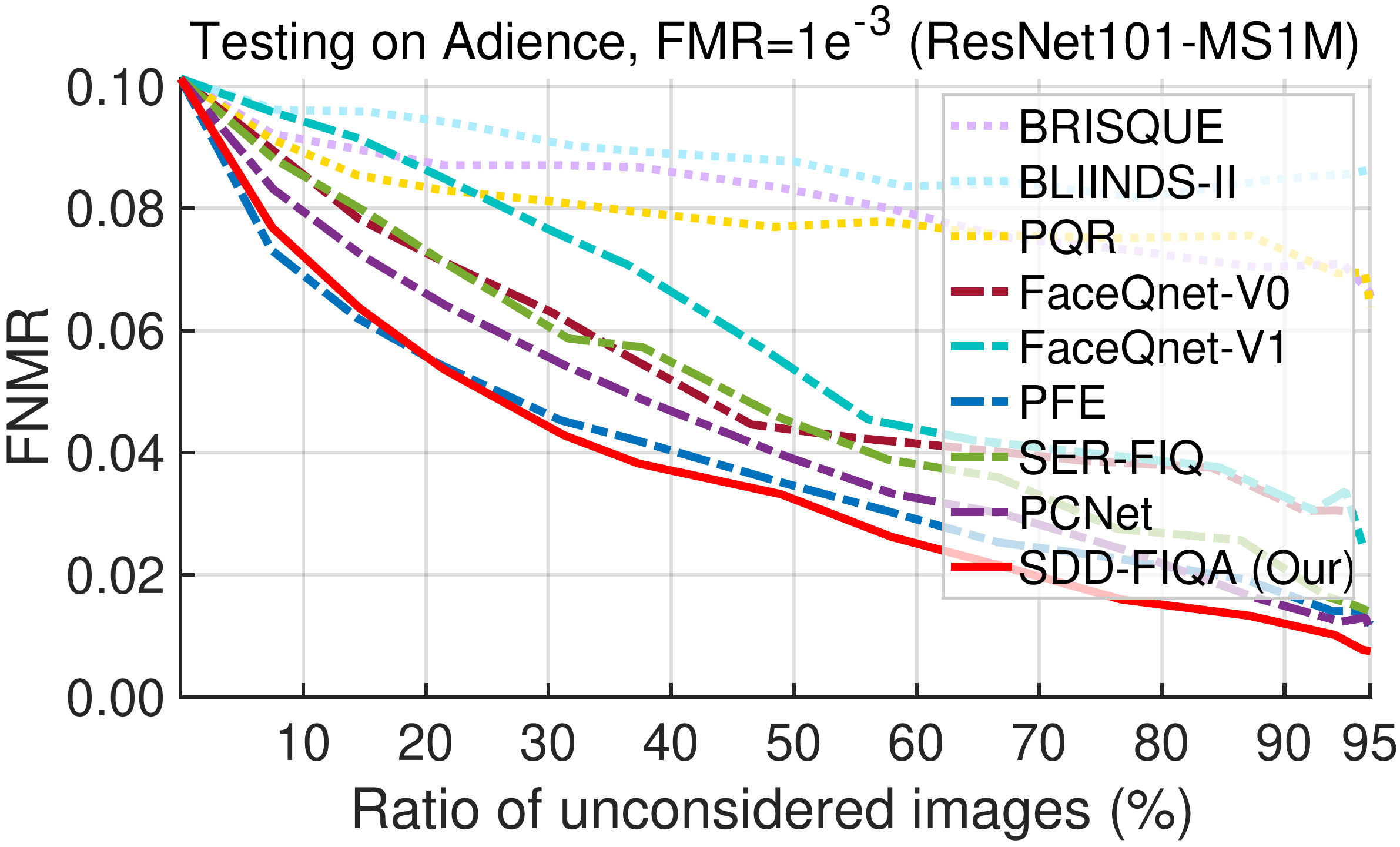}
 }\hspace{-6mm}
 \quad
 \subfigure[]{
 \includegraphics[width=5.75cm]{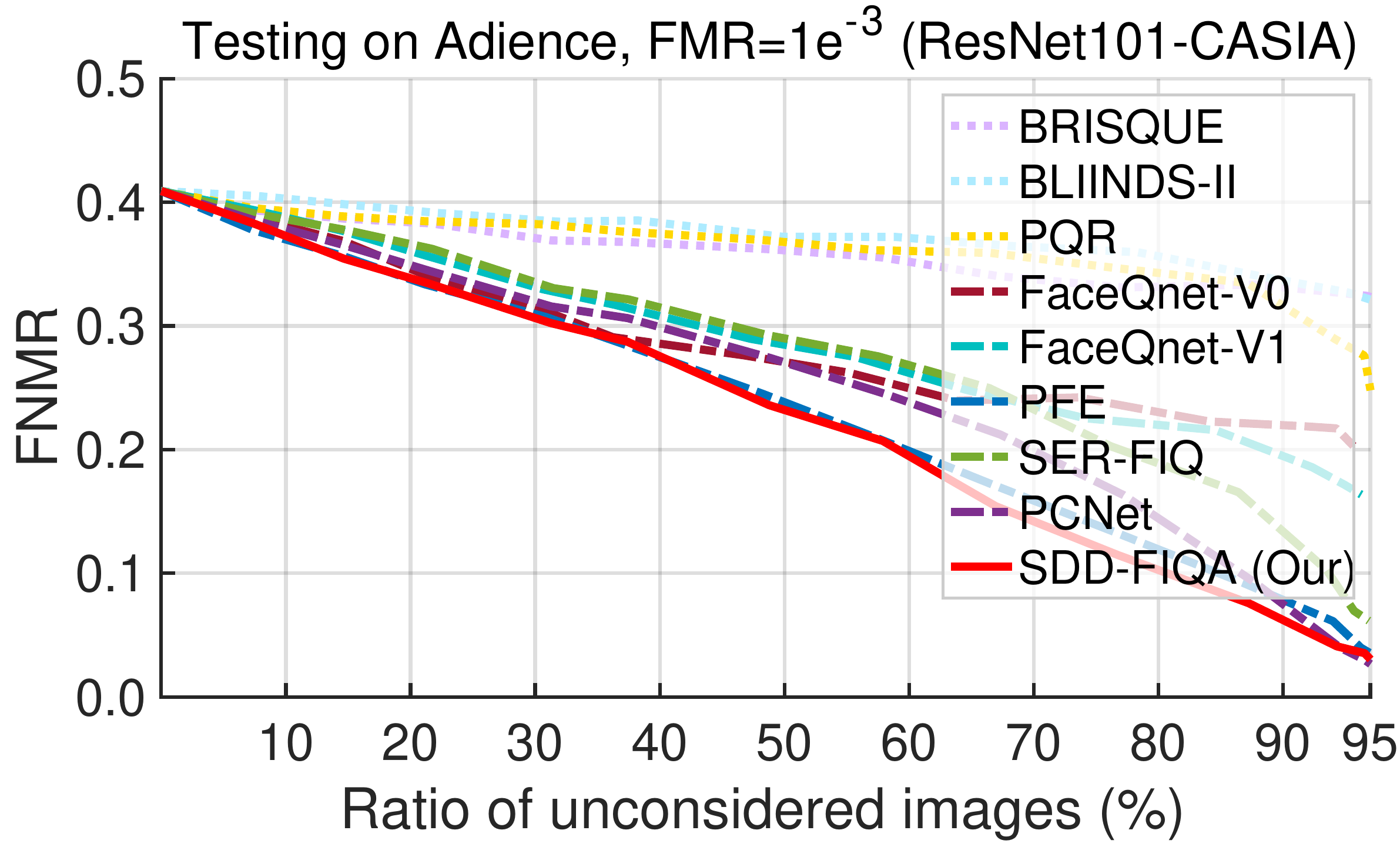}
 }\hspace{-6mm}
 \quad
 \subfigure[]{
 \includegraphics[width=5.75cm]{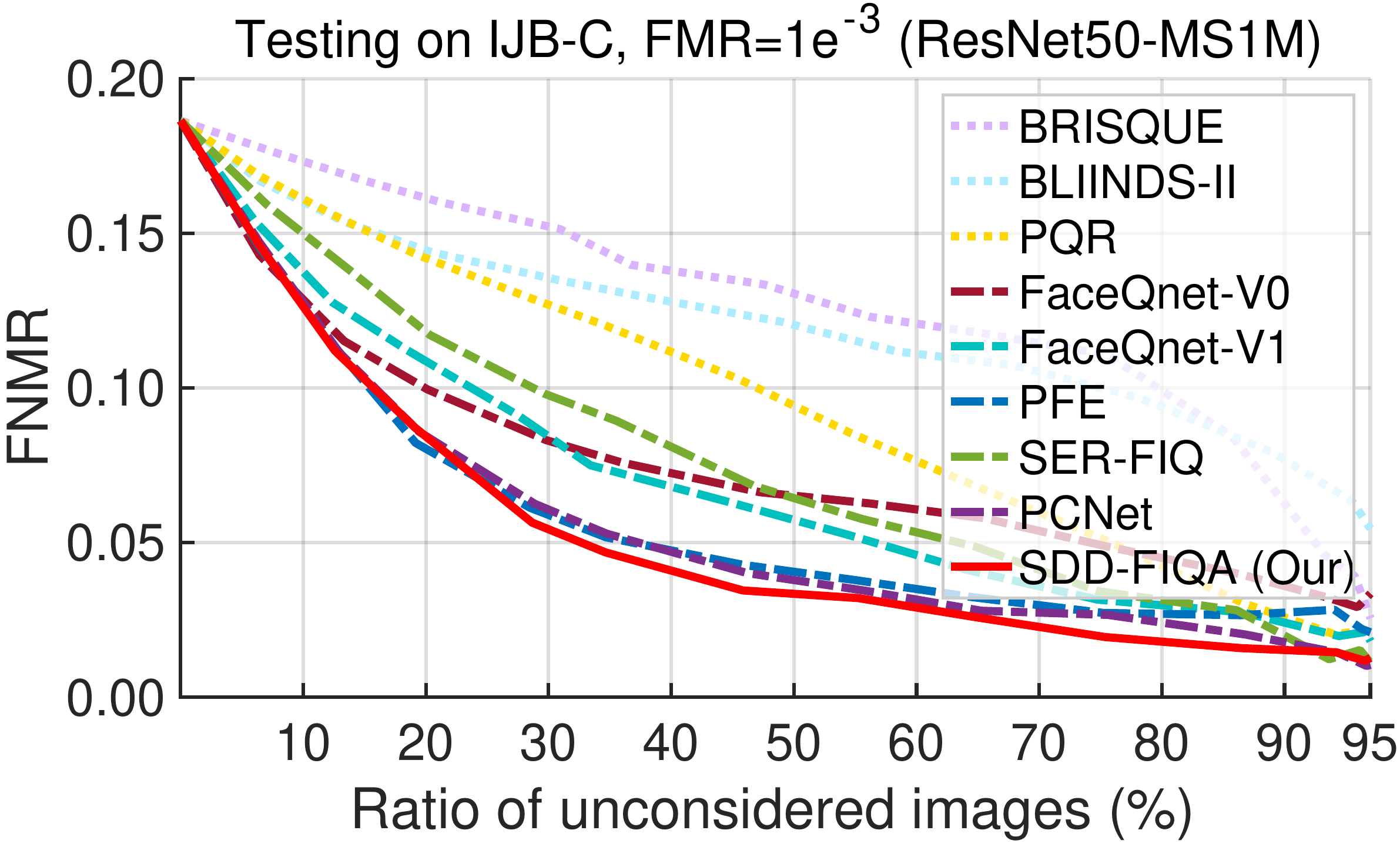}
 }\hspace{-6mm}
 \quad
 \subfigure[]{
 \includegraphics[width=5.75cm]{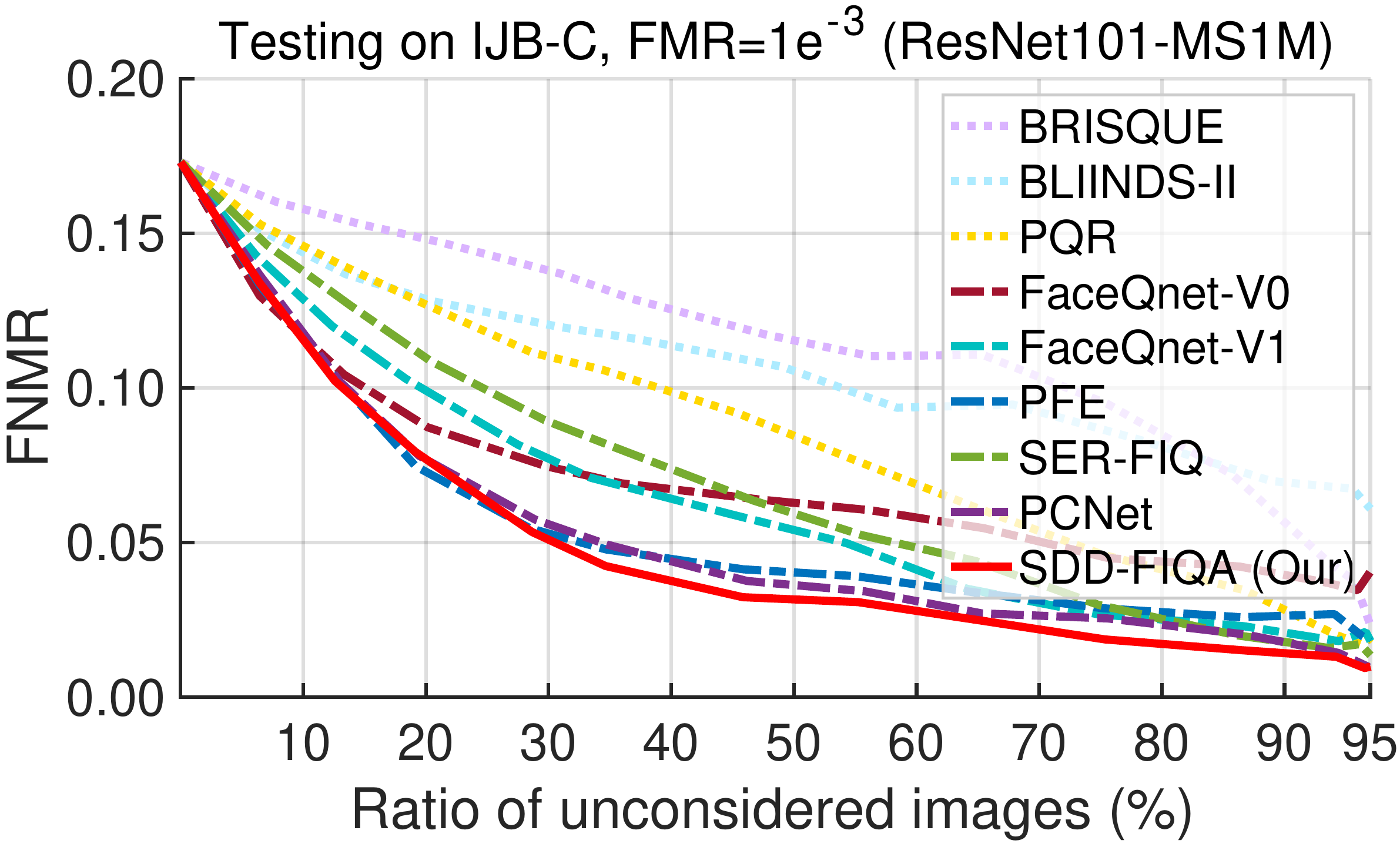}
 }\hspace{-6mm}
 \quad
 \subfigure[]{
 \includegraphics[width=5.75cm]{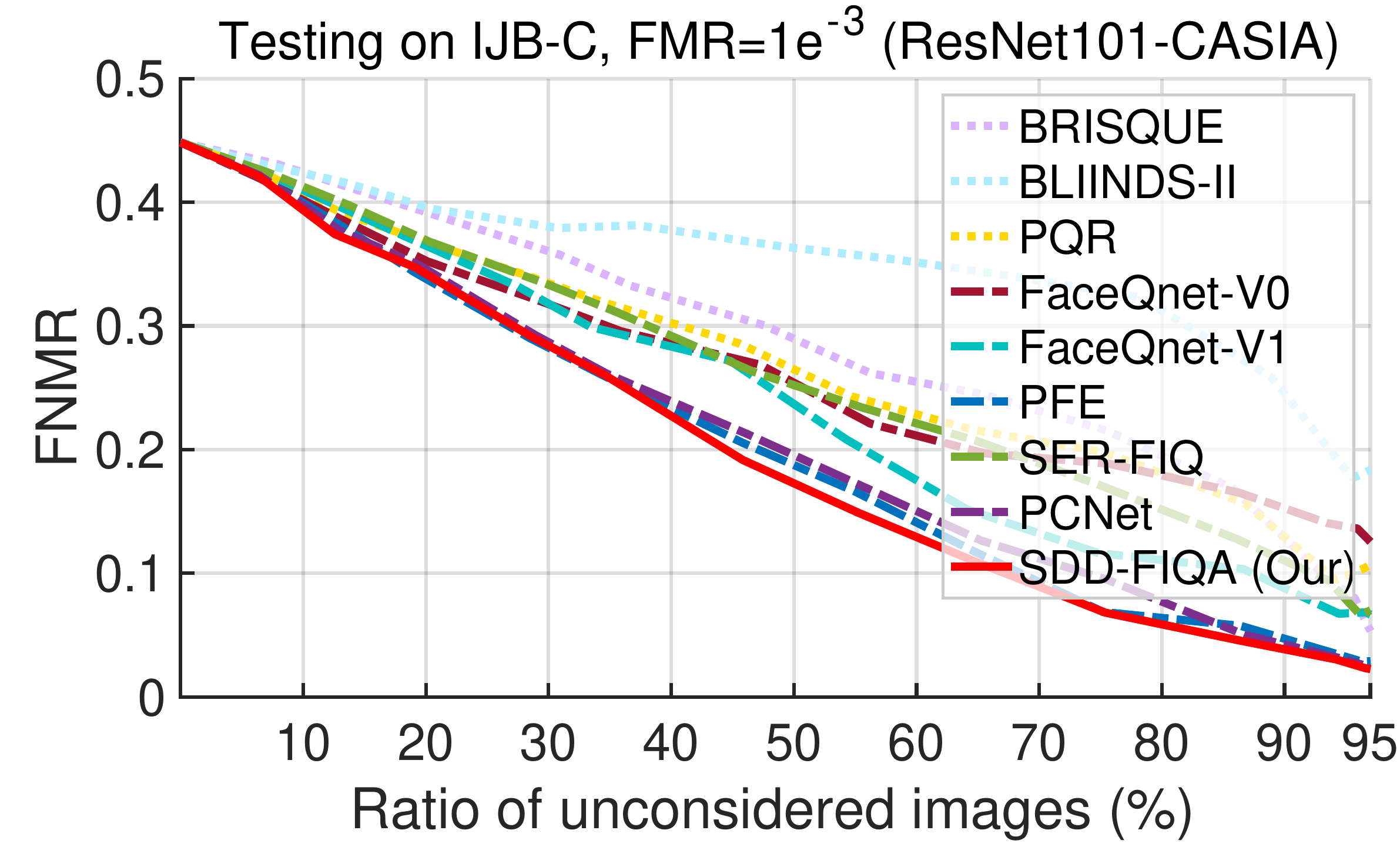}
 }
 \caption{Face verification performance on the predicted face image quality scores. The EVRC shows the effectiveness of rejecting low-quality face images in terms of FNMR at a threshold of $1e^{-3}$ FMR.}\label{EVRC}
 \end{figure*}
\noindent{\textbf{Face recognition models}}. Different face recognition models are employed, containing ResNet50 trained on MS1M (ResNet50-MS1M), ResNet101 trained on MS1M (ResNet101-MS1M),  and ResNet101 trained on CASIA (ResNet101-CASIA). All recognition models are learned by ArcFace with 512-dimensional embedding features \cite{ArcFace}. 

\noindent{\textbf{Implementation details}}. Our network is built on the PyTorch framework and on a machine equipped with eight NVIDIA Tesla P40 GPUs. Face images are all aligned, scaled, and cropped to 112 $\times$ 112 pixels by MTCNN \cite{MTCNN}. During the training stage, all networks are learned with Adam optimizer with weight decay $5e^{-4}$. The initial learning rate is $1e^{-3}$ and drops by a factor of $5e^{-2}$ every 5 epochs. We set $m=24$ and $K=12$ in Eq.~\eqref{eq3.6} and the parameter $\zeta$ is set to 1 in Eq. \eqref{huber}.

\noindent{\textbf{Evaluation protocols}}. As done in  \cite{Best2018,faceqnetv02019,faceqnetv12020,SER-FIQ2020,Schlett2020}, we use EVRC to evaluate the performance of FIQA methods . Moreover, we introduce Area Over Curve (AOC) to quantify the EVRC results, which is defined by
\begin{equation}\label{AOC}
    \mathrm{AOC}=1-\int_{a}^{b} g(\varphi) d\varphi,
\end{equation}
where $g(\varphi)$ denotes the FNMR at $\varphi$, $\varphi=1-\sigma$ is the ratio of unconsidered images, $a$ and $b$ denote lower and upper bounds and are set to be 0 and 0.95 in our experiments, respectively.

\subsection{Results \& Analysis}
The proposed SDD-FIQA method is compared with eight state-of-the-arts covering analytics-based methods (\emph{e.g.}  BRISQUE \cite{BRISQUE}, BLIINDS-II \cite{BLIINDS-II}, and PQR \cite{PQR}) and learning-based FIQA methods (\emph{e.g.} FaceQnet-V0 \cite{faceqnetv02019},  FaceQnet-V1 \cite{faceqnetv12020}\footnote{https://github.com/uam-biometrics/FaceQnet.}, PFE \cite{PFE2019}\footnote{https://github.com/seasonSH/Probabilistic-Face-Embeddings.}, SER-FIQ (same model)\footnote{https://github.com/pterhoer/FaceImageQualit.}, \cite{SER-FIQ2020}, and PCNet \cite{PCNet2020}). All the compared methods are reproduced following their paper settings or using their released code directly. In the following, we compare the proposed SDD-FIQA  with competitors under the same recognition model setting and the cross recognition model setting to verify the FIQA performance and generalization. We carry out comprehensive experiments to justify the superiority of  our SDD-FIQA in terms of the quality weighting improvement of set-based face recognition verification \cite{PCNet2020}, unbiasedness of bias factors \cite{DBLP}. Finally, we also implement ablation study to demonstrate the effectiveness of each key  components of our method.


\subsubsection{Same Recognition Model Performance}\label{4.2.1}
In this part, we follow \cite{SER-FIQ2020} to evaluate the performance under the same recognition model. ResNet50-MS1M is selected to generate quality pseudo-labels and training quality regression network. As shown in Fig. \ref{EVRC}, the FNMR of SDD-FIQA sharply decreases with the increase of the ratio of unconsidered images and outperforms all the compared methods by a large margin. Moreover, we also report the AOC results in Tab. \ref{ACC_Res50_MS1M}. The results show that our SDD-FIQA improves the best competitor by 13.9\% for LFW, 5.2\% for Adience, and 2.4\% for IJB-C in terms of the average (Avg) performance.

%
%

\subsubsection{Cross Recognition Model Performance}\label{4.2.2}
In real-world applications, the recognition models for training and testing may not be the same. To verify the generalization of the proposed SDD-FIQA, we perform the cross recognition model experiment with two settings: 1) First employ ResNet50-MS1M for generating quality pseudo-labels and training quality regression network, and then adopt ResNet101-MS1M as the deployed recognition model for testing; 2) First employ ResNet50-MS1M for generating quality pseudo-labels and training quality regression network, and then adopt ResNet101-CASIA as the deployed recognition model for testing. To the best of our knowledge,  we are the first to evaluate the generalization of FIQA methods.

\begin{table}
\setlength{\abovecaptionskip}{0.05cm}
\setlength{\belowcaptionskip}{-0.1cm}
\begin{center}
\caption{The set-based verification TAR results on IJB-C. The deployed recognition model is ResNet50-MS1M.}\label{Set_based}
\resizebox{236pt}{25pt}{
\begin{tabular}{|c|c|c|c|c|}
\Xhline{0.6pt}
\bf{Weighting}								&\bf{FMR=}$\bm{1e^{-2}}$ 	&\bf{FMR=}$\bm{1e^{-3}}$		&\bf{FMR=}$\bm{1e^{-4}}$   &\bf{Avg}\\
\Xhline{0.6pt}
Baseline 					                &0.9715   &0.9512   &0.9191   &0.9473  \\ 
SER-FIQ \cite{SER-FIQ2020} 			&0.9755   &0.9559   &0.9318   &0.9544  \\ 
PCNet \cite{PCNet2020} 				&0.9766   &0.9602   &0.9461   &0.9610   \\ 
\Xhline{0.6pt}
\bf{SDD-FIQA (Our)} 		&\bf{0.9773}   &\bf{0.9605}   &\bf{0.9473}	&\bf{0.9617}   \\ 
\Xhline{0.6pt}
\end{tabular}}
\end{center}
\end{table}

\begin{figure}
\centering
\subfigure[]{
\includegraphics[width=2.71cm]{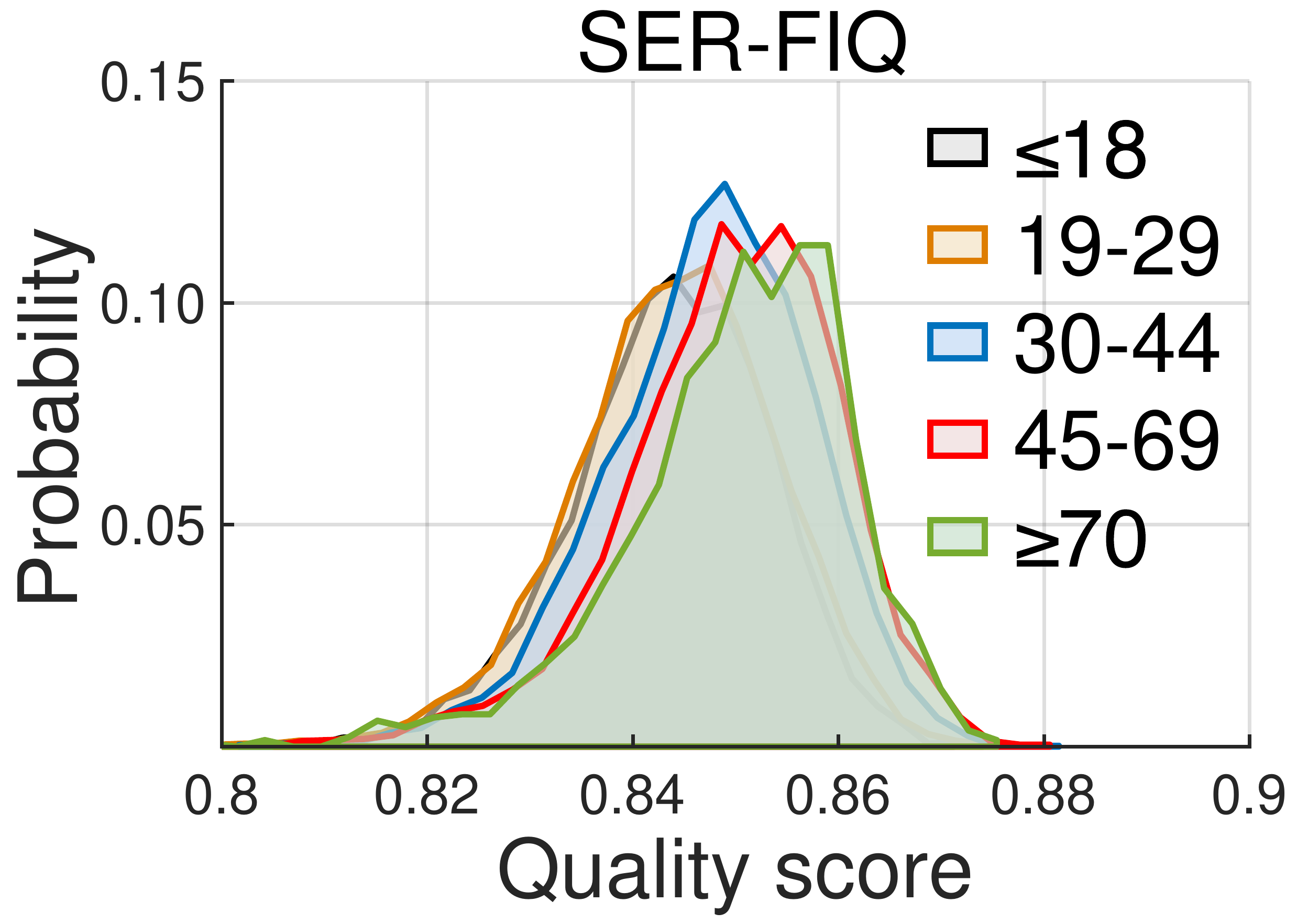}
}\hspace{-6mm}
\quad
\subfigure[]{
\includegraphics[width=2.71cm]{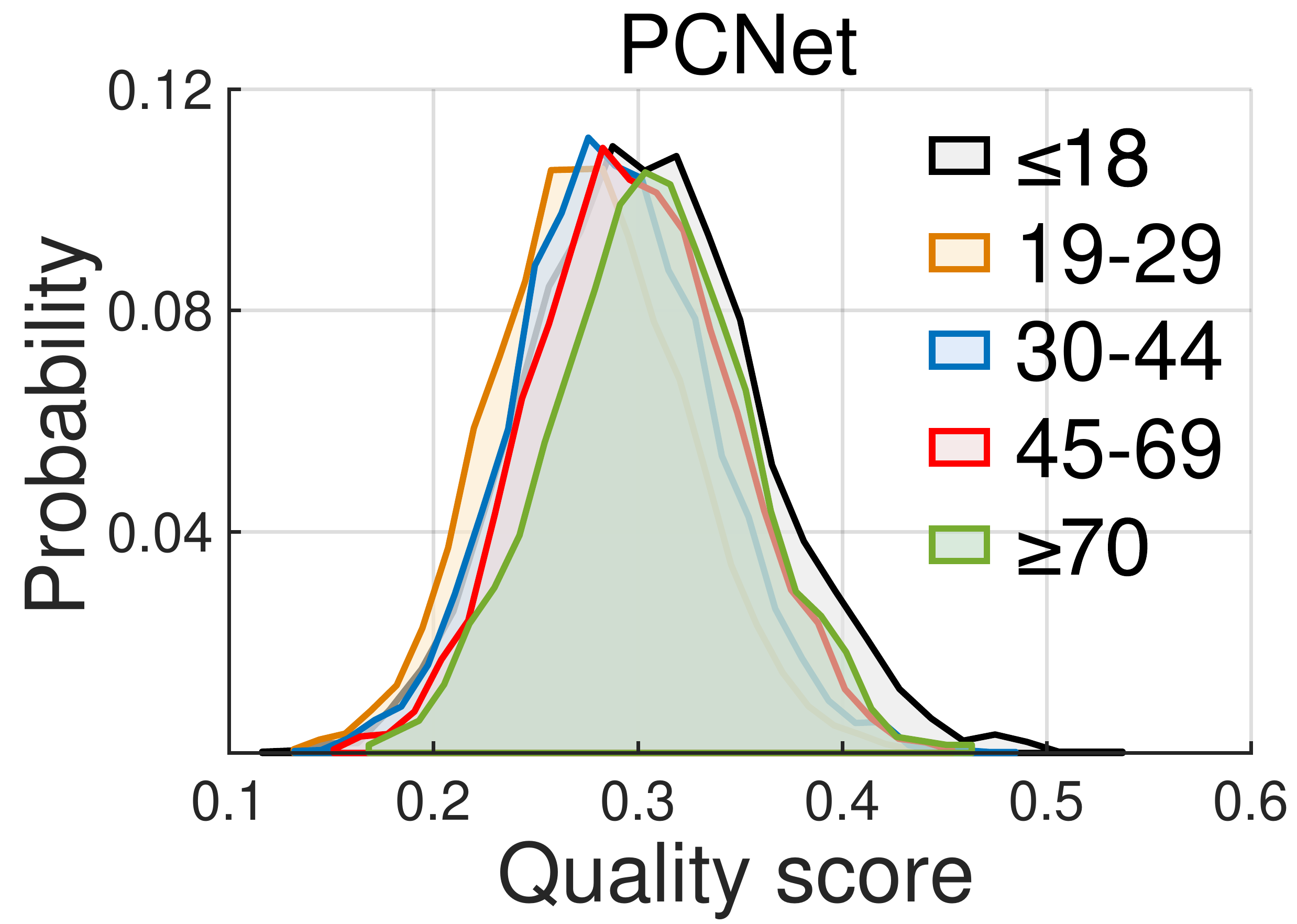}
}\hspace{-6mm}
\quad
\subfigure[]{
\includegraphics[width=2.71cm]{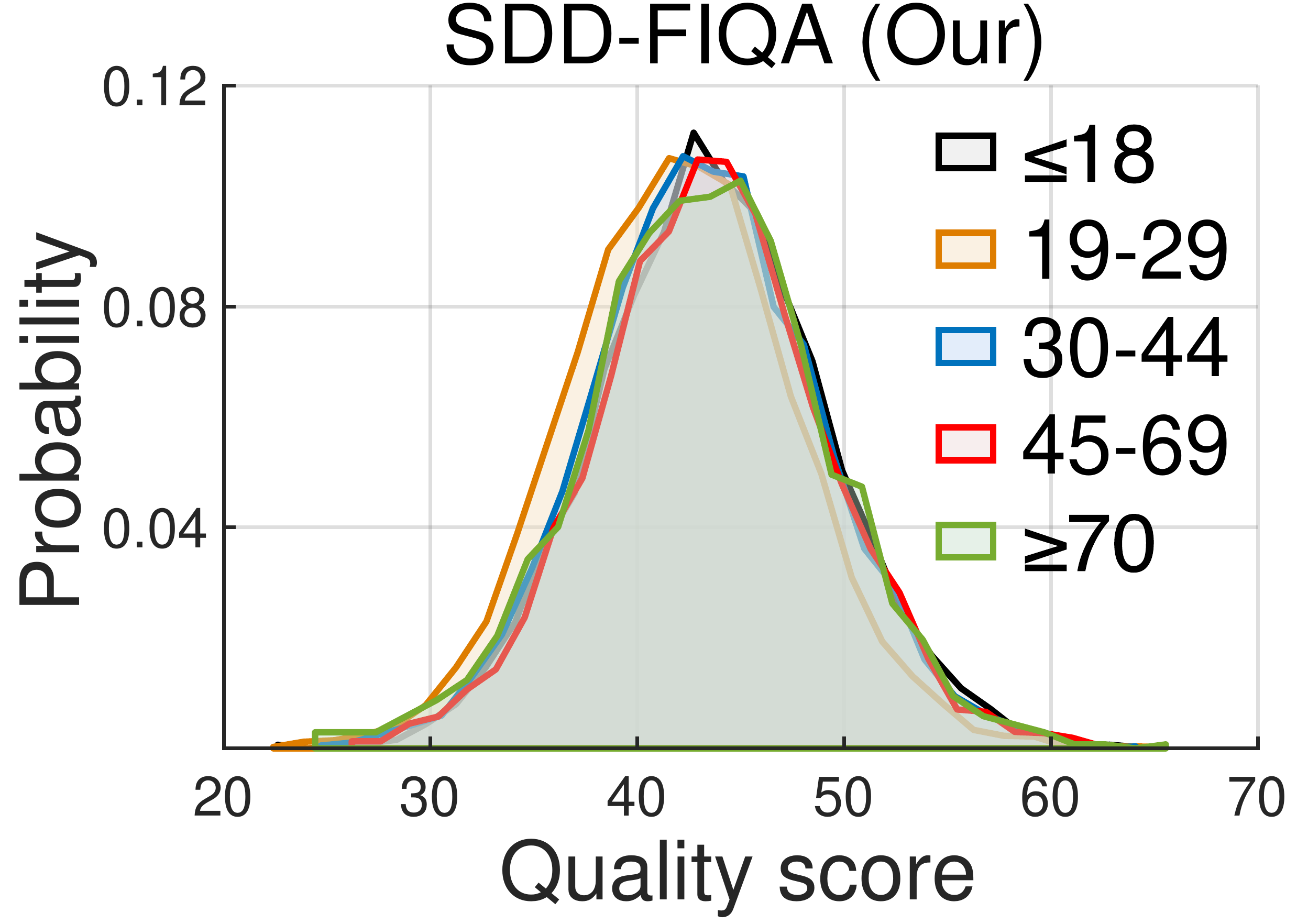}
}\hspace{-6mm}
\quad
\subfigure[]{
\includegraphics[width=2.71cm]{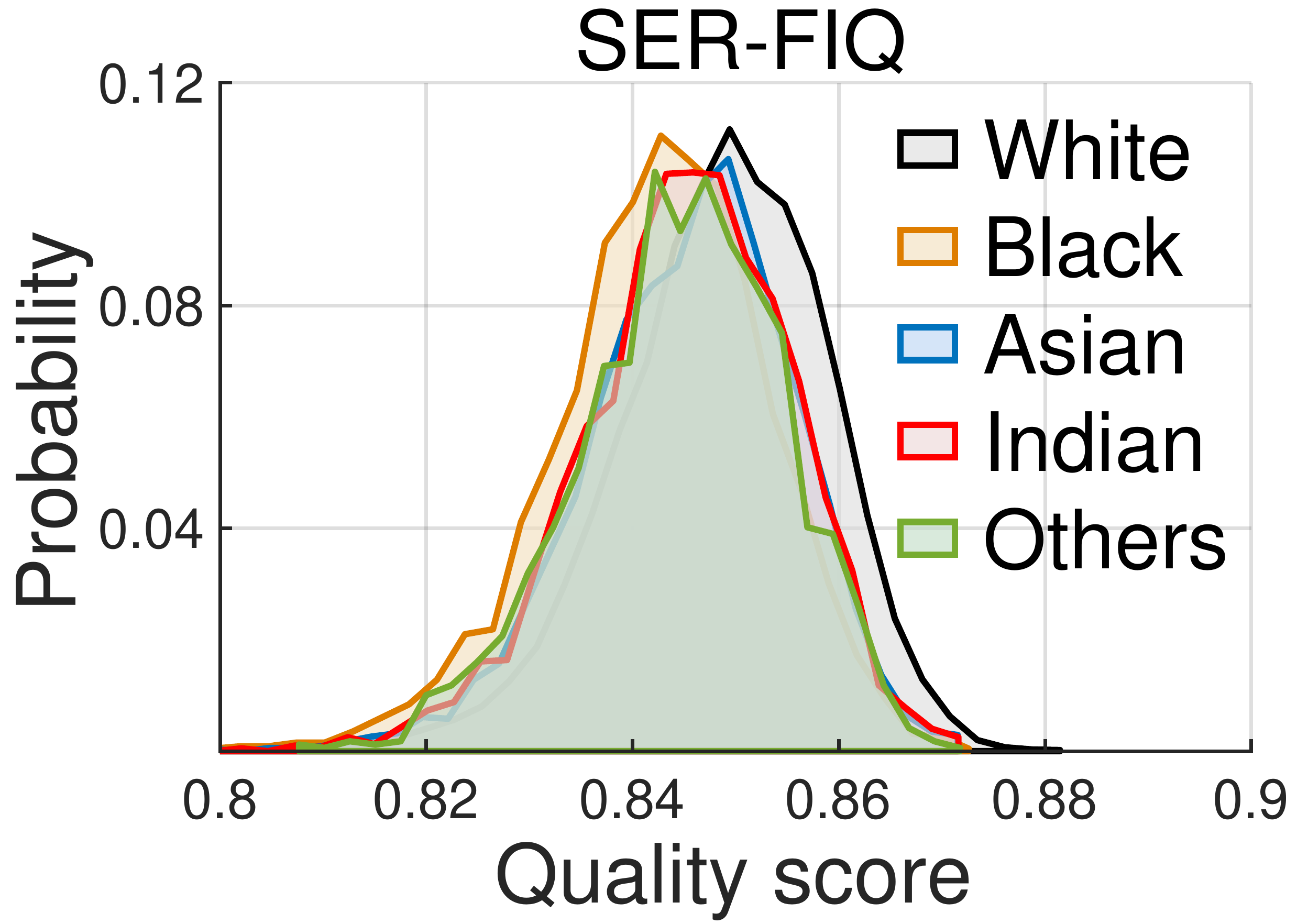}
}\hspace{-6mm}
\quad
\subfigure[]{
\includegraphics[width=2.71cm]{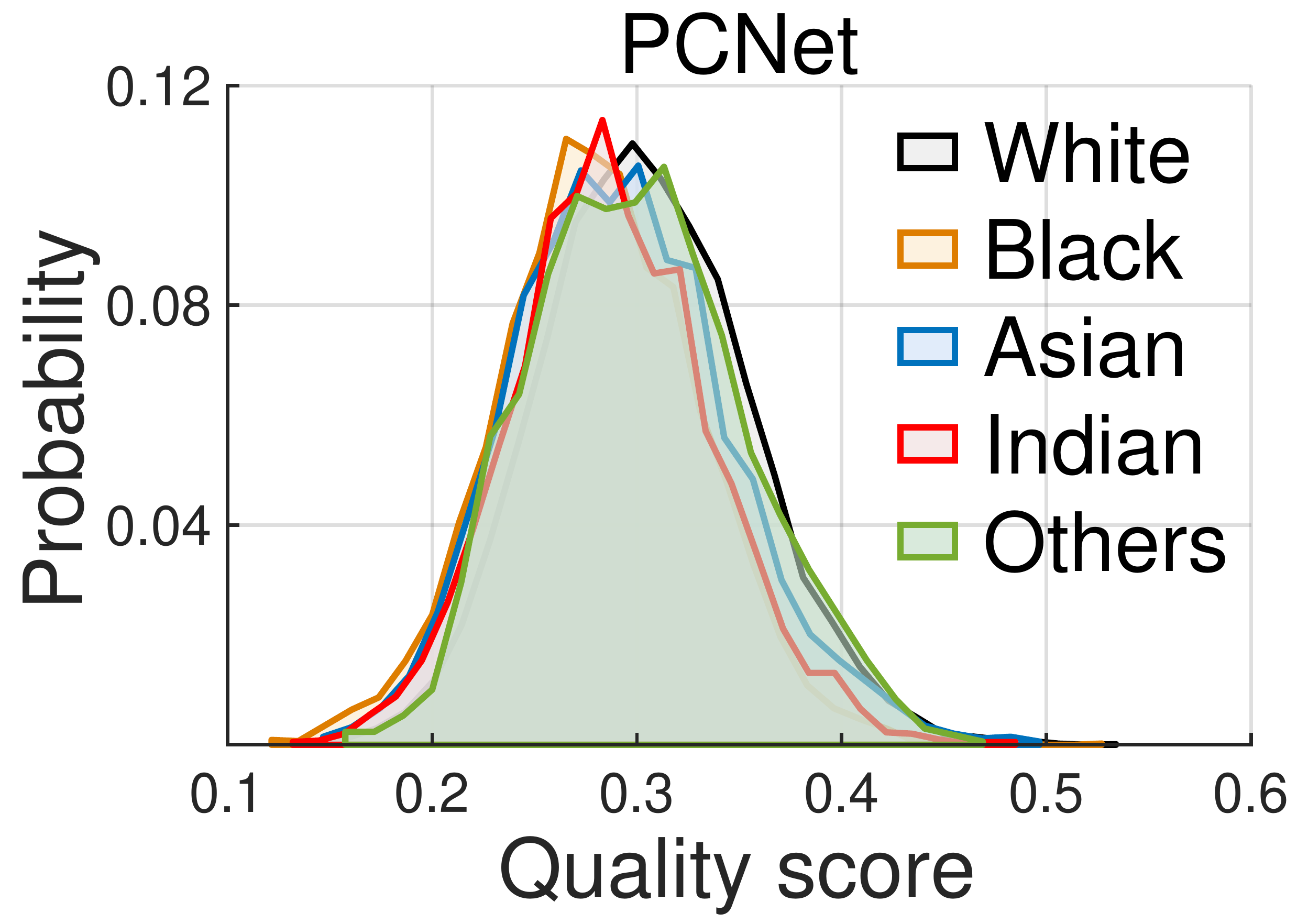}
}\hspace{-6mm}
\quad
\subfigure[]{
\includegraphics[width=2.71cm]{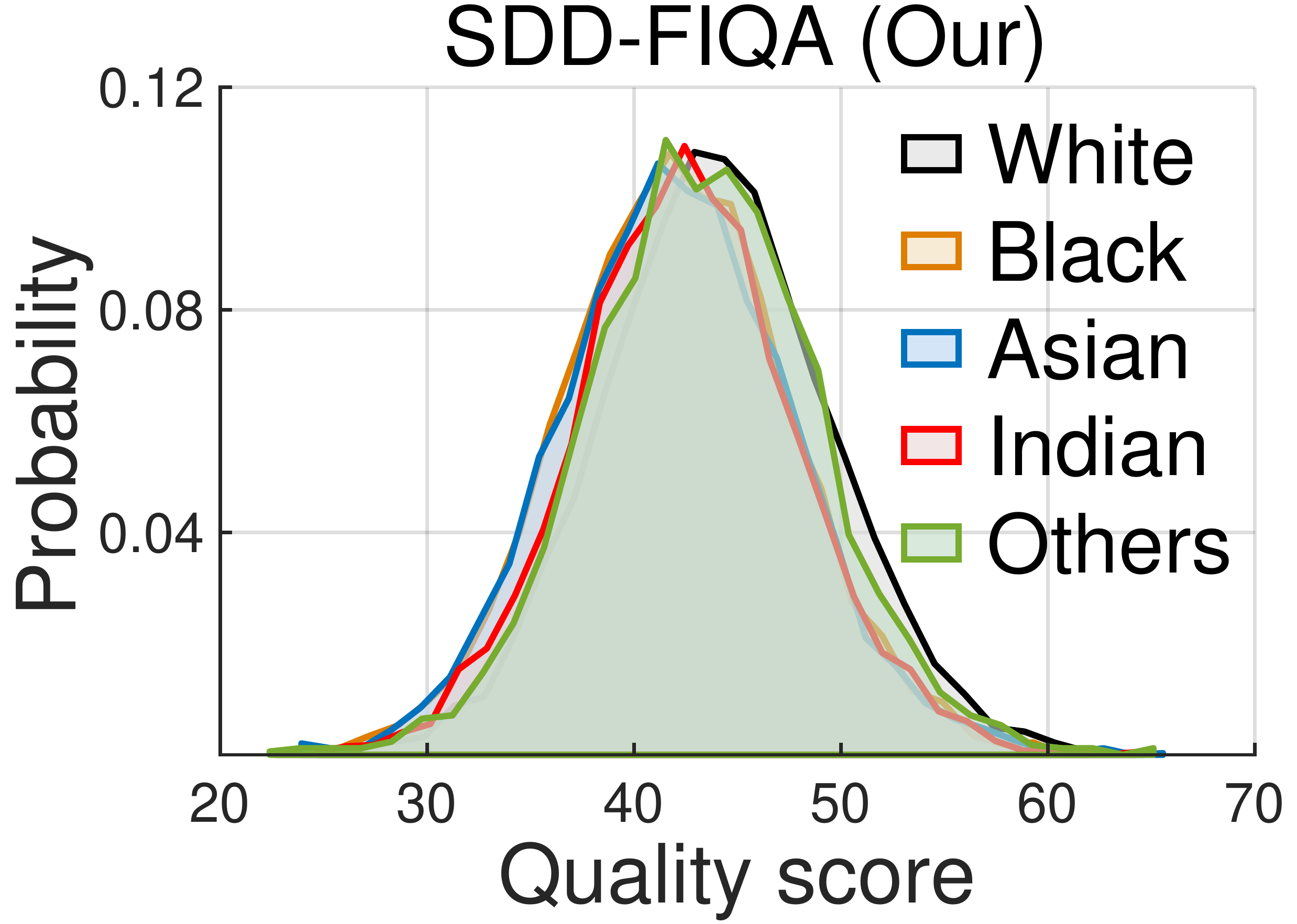}
}
\caption{The distributions of quality scores with different bias factors. The first row is for age factor and the second row is for race factor.}
\label{Bias}
\end{figure}

The EVRC results are shown in Fig. \ref{EVRC} and the AOC results are reported in Tab. \ref{ACC_Res101_MS1M}. Fig. \ref{EVRC} (b), (e) and (h) de- monstrate the results on the first cross model setting and Fig. \ref{EVRC} (c), (f) and (i) show the results on the second cross model setting. We can see that SDD-FIQA performs better than all the compared methods on LFW, Adience, and IJB-C. Moreover, The gap is enlarged with the increase of the ratio of unconsidered images. The results in Tab. \ref{ACC_Res101_MS1M} also show that SDD-FIQA achieves over 17.5\% on LFW, 6.5\% on Adience, and 2.7\% on IJB-C higher accuracy than the best competitor in terms of the Avg AOC performance. Tab. \ref{ACC_Res101_CASIA} demonstrates that SDD-FIQA outperforms the best competitor by 8.3\% on LFW,  3.7\% on Adience, and 1.1\% on IJB-C, respectively. The above results demonstrate that the proposed SDD-FIQA has better generalization than counterparts.

\subsubsection{Set-based Verification with Quality Weighting}\label{4.2.3}
We perform the set-based verification on IJB-C dataset to verify the face quality impact in terms of the recognition performance. Specifically, as described in \cite{PCNet2020}, similarities are collected by set-to-set mapping, in which each set consists of a variable number of images or video frames from different sources. Then the set descriptor is calculated as a weighted average of individual faces by  $f(x)=\frac{1}{n}\sum_{i=1}^{n} \phi_{\pi}(x_i) \cdot f(x_i)$, where $x$ denotes the identity corresponding to $x_i$. We evaluate the True Accept Rate (TAR) scores under different FMR thresholds following \cite{PCNet2020}. The set-based verification results are reported in Tab. \ref{Set_based}. The results show that the quality weight of our method is more beneficial to verification performance.

\begin{table}
\begin{center}
\caption{AOC results on cross recognition model setting. The deployed recognition model is ResNet101-MS1M.}\label{ACC_Res101_MS1M}
\resizebox{236pt}{135pt}{
\begin{tabular}{|C{0.7cm}|c|c|c|c|c|}
\Xhline{0.6pt}
$ $	 &\bf{LFW}	&\bf{FMR}=$\bm{1e^{-2}}$ 	&\bf{FMR=}$\bm{1e^{-3}}$		&\bf{FMR=}$\bm{1e^{-4}}$ 	&\bf{Avg}\\
\Xhline{0.6pt}
\multirow{3}*{\rotatebox{90}{\makecell{\textbf{\small{Analytics}}\\ \textbf{\small{Based}}}}}
& BRISQUE \cite{BRISQUE}				&0.0700   &0.1200   &0.1779   &0.1227 \\
& BLIINDS-II \cite{BLIINDS-II}		&0.2035   &0.2004   &0.2056   &0.2032 \\
& PQR \cite{PQR}						&0.3508   &0.2657   &0.2995   &0.3053 \\
\hline
\multirow{5}*{\rotatebox{90}{\makecell{\textbf{\small{Learning}}\\ \textbf{\small{Based}}}}}
& FaceQnet-V0 \cite{faceqnetv02019} 	&0.5277   &0.5757   &0.5707   &0.5580 \\ 
& FaceQnet-V1 \cite{faceqnetv12020} 	&0.5002   &0.5158   &0.5901   &0.5354 \\
& PFE \cite{PFE2019}					&0.5402   &0.5587   &0.5828   &0.5606 \\ 
& SER-FIQ \cite{SER-FIQ2020} 		&0.6027   &0.6401   &0.7011   &0.6480 \\
& PCNet \cite{PCNet2020} 		&0.6774   &0.6915   &0.6681   &0.6790 \\  
\Xhline{0.6pt}
& \bf{SDD-FIQA (Our)} 						&\bf{0.8181}   &\bf{0.7881}   &\bf{0.7874}   &\bf{0.7979} \\ 
\hline
\hline
\hline
$ $	 &\bf{Adience}	&\bf{FMR}=$\bm{1e^{-2}}$ 	&\bf{FMR=}$\bm{1e^{-3}}$		&\bf{FMR=}$\bm{1e^{-4}}$ 	&\bf{Avg}\\
\Xhline{0.6pt}
\multirow{3}*{\rotatebox{90}{\makecell{\textbf{\small{Analytics}}\\ \textbf{\small{Based}}}}}
& BRISQUE \cite{BRISQUE}				&0.2773   &0.2097   &0.2412   &0.2428 \\
& BLIINDS-II \cite{BLIINDS-II}		&0.1425   &0.1416   &0.2660   &0.1834 \\
& PQR \cite{PQR}						&0.2697   &0.2322   &0.2601   &0.2540 \\
\hline
\multirow{5}*{\rotatebox{90}{\makecell{\textbf{\small{Learning}}\\ \textbf{\small{Based}}}}}
& FaceQnet-V0 \cite{faceqnetv02019} 	&0.4380   &0.4874   &0.4837   &0.4697 \\ 
& FaceQnet-V1 \cite{faceqnetv12020} 	&0.3475   &0.4196   &0.4721   &0.4131 \\
& PFE \cite{PFE2019}					&0.5130   &0.6168   &0.5909   &0.5736 \\
& SER-FIQ \cite{SER-FIQ2020}  		&0.4763   &0.5163   &0.4768   &0.4898 \\ 
& PCNet \cite{PCNet2020} 		&0.4959  &0.5694   &0.5630   &0.5428 \\  
\Xhline{0.6pt}
& \bf{SDD-FIQA (Our)} 						&\bf{0.5563}   &\bf{0.6415}   &\bf{0.6365}   &\bf{0.6114} \\ 
\hline
\hline
\hline
$ $	 &\bf{IJB-C}	&\bf{FMR}=$\bm{1e^{-2}}$ 	&\bf{FMR=}$\bm{1e^{-3}}$		&\bf{FMR=}$\bm{1e^{-4}}$ 	&\bf{Avg}\\
\Xhline{0.6pt}
\multirow{3}*{\rotatebox{90}{\makecell{\textbf{\small{Analytics}}\\ \textbf{\small{Based}}}}}
& BRISQUE \cite{BRISQUE}				&0.2745   &0.3461   &0.4003   &0.3403 \\
& BLIINDS-II \cite{BLIINDS-II}		&0.3819   &0.3915  &0.4074  &0.3936\\
& PQR \cite{PQR}						&0.5124   &0.5110   &0.5485   &0.5240 \\
\hline
\multirow{5}*{\rotatebox{90}{\makecell{\textbf{\small{Learning}}\\ \textbf{\small{Based}}}}}
& FaceQnet-V0 \cite{faceqnetv02019} 	&0.6240   &0.6022   &0.6445   &0.6236 \\ 
& FaceQnet-V1 \cite{faceqnetv12020} 	&0.6399  &0.6376   &0.6455   &0.6410 \\
& PFE \cite{PFE2019}					&0.6790   &0.6910   &0.7239   &0.6980 \\
& SER-FIQ \cite{SER-FIQ2020}  		&0.6134   &0.6044   &0.6473   &0.6217 \\ 
& PCNet \cite{PCNet2020} 		&0.7016   &0.7031   &0.7280   &0.7109 \\ 
\Xhline{0.6pt}
& \bf{SDD-FIQA (Our)} 						&\bf{0.7160}   &\bf{0.7243}   &\bf{0.7500}   &\bf{0.7301} \\ 
\Xhline{0.6pt}
\end{tabular}}
\end{center}
\end{table}

 \subsubsection{\textbf{Investigation of Bias Factors}}\label{4.2.5}
 As done in \cite{DBLP}, we also implement an experiment to investigate whether there is a significant correlation between the quality score and the bias factors such as age and race. The investigation results on the UTKFace dataset are shown in Fig. \ref{Bias}. It is known that the greater the distance between the different bias factor distributions is, the stronger correlation between the predicted quality scores and the face image bias factors. We can find that the quality scores predicted by our SDD-FIQA have a smaller bias than the SER-FIQ and PCNet methods in terms of the age and race factors. This investigation further proves that our SDD-FIQA is more reliable and robust than the stat-of-the-arts.

 \subsubsection{\textbf{Ablation Study}}\label{4.2.4}
\begin{figure}
\centering
\subfigure[]{
\includegraphics[width=3.8cm]{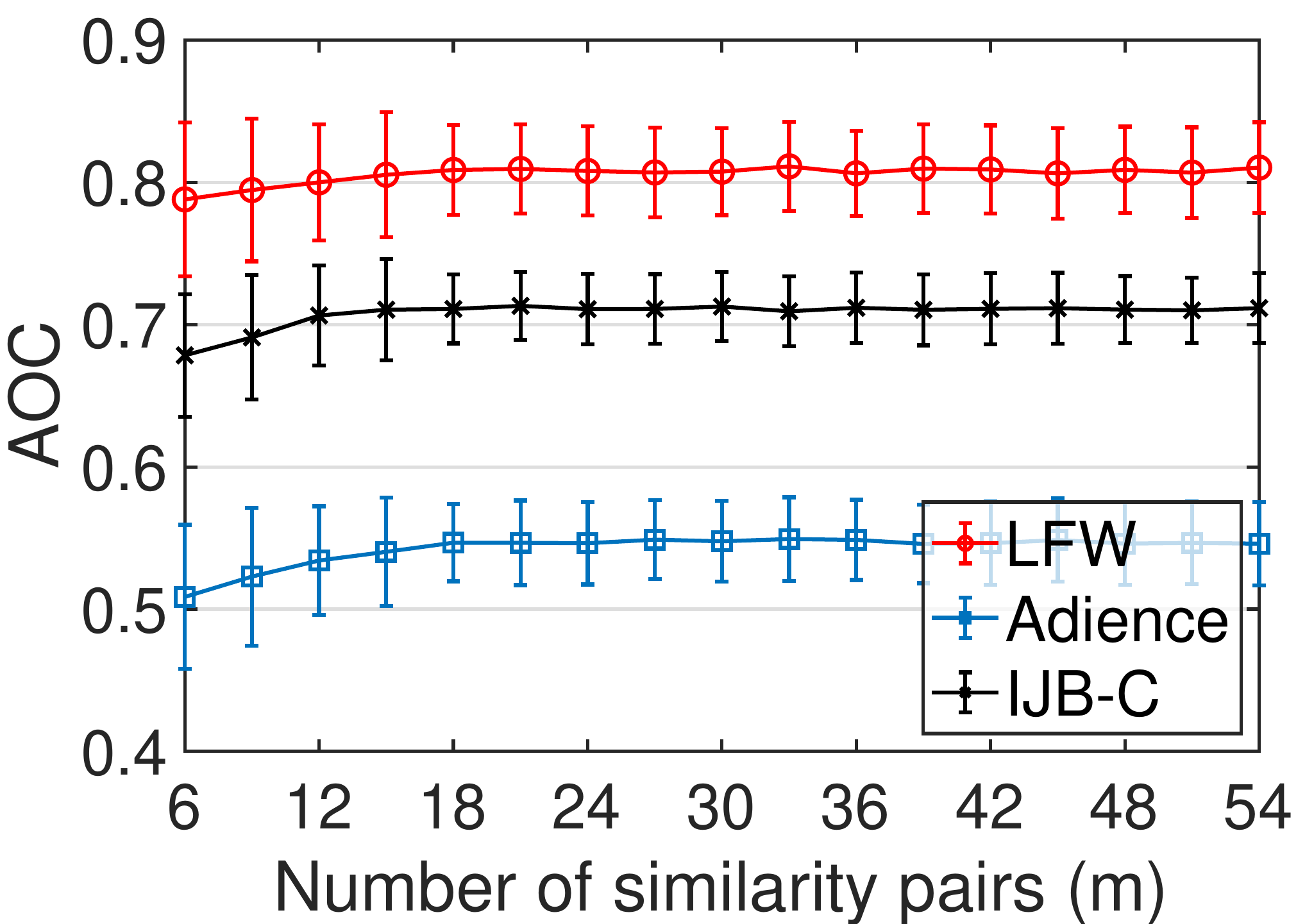}
}\hspace{-5mm}
\quad
\subfigure[]{
\includegraphics[width=3.8cm]{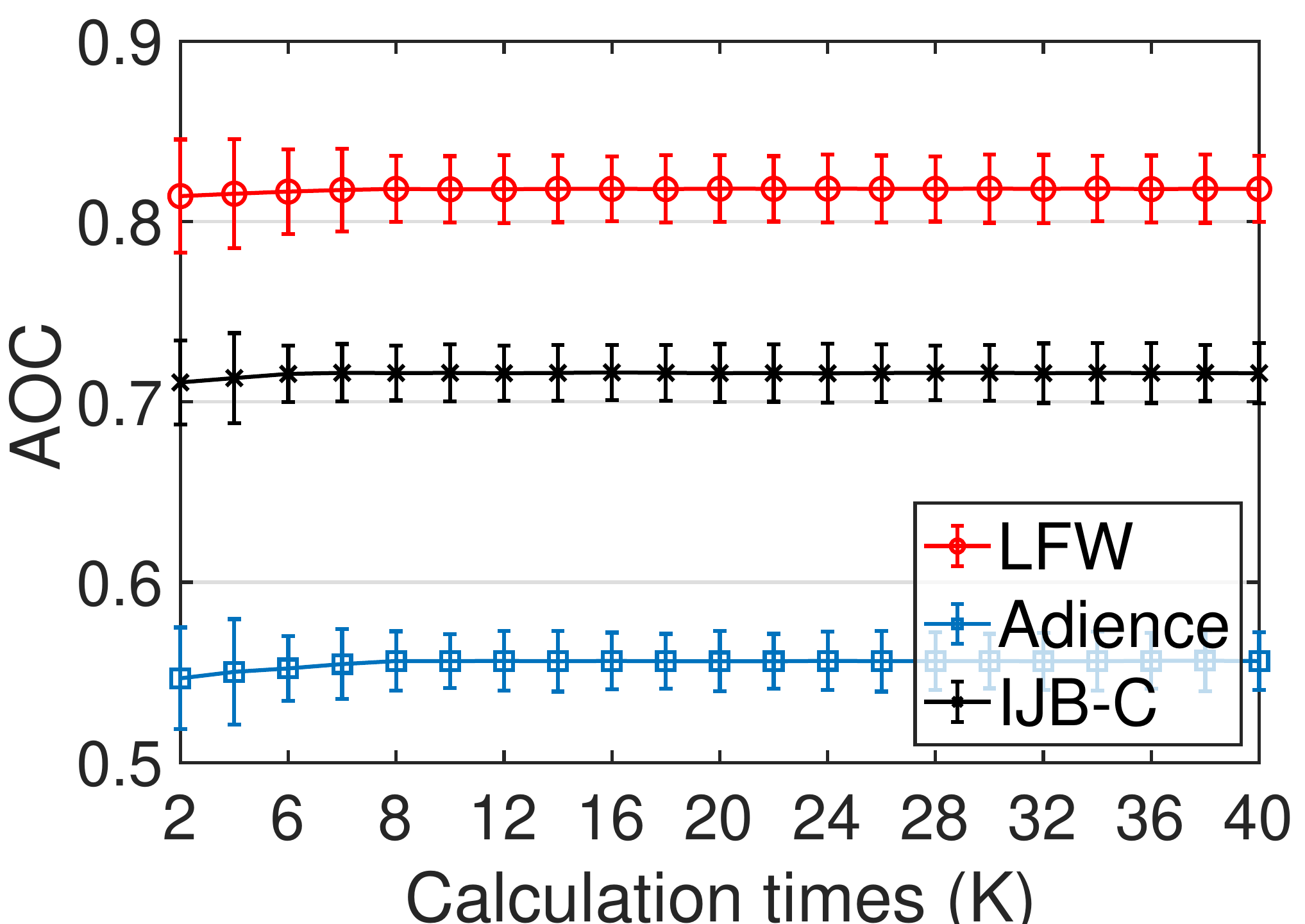}
}
\caption {Empirical studies on the parametric sensitivity of SDD-FIQA. (a) is for number of similarity pairs $m$. (b) is fo calculation times $K$.}
\label{mandK}
\end{figure}

\noindent{\textbf{Parametric sensitivity}}.  
To confirm the parametric sensitivity of SDD-FIQA, we evaluate the AOC results with different hyper-parameters on the three benchmark datasets, which is shown in Fig. \ref{mandK}. To be specific, we fix $K=1$ and vary $m$ from 6 to 54 to select a proper value of $m$ in the first experiment.  Then we fix $m = 24$ and vary $K$ from 2 to 40 to determine the suitable value of $K$ in the second experiment. All experiments are performed on ResNet101-MSIM as the deployed recognition model under FMR=$1e^{-2}$. We can notice that with the increases of $K$ and $m$, the AOC stably increases and converges until 24 for the number of similarity pairs and 12 for calculation times.  Thus in our experiments, $m$ and $K$ are set to 24 and 12, respectively.\par

\begin{figure}
\centering
\includegraphics[width=6.4cm]{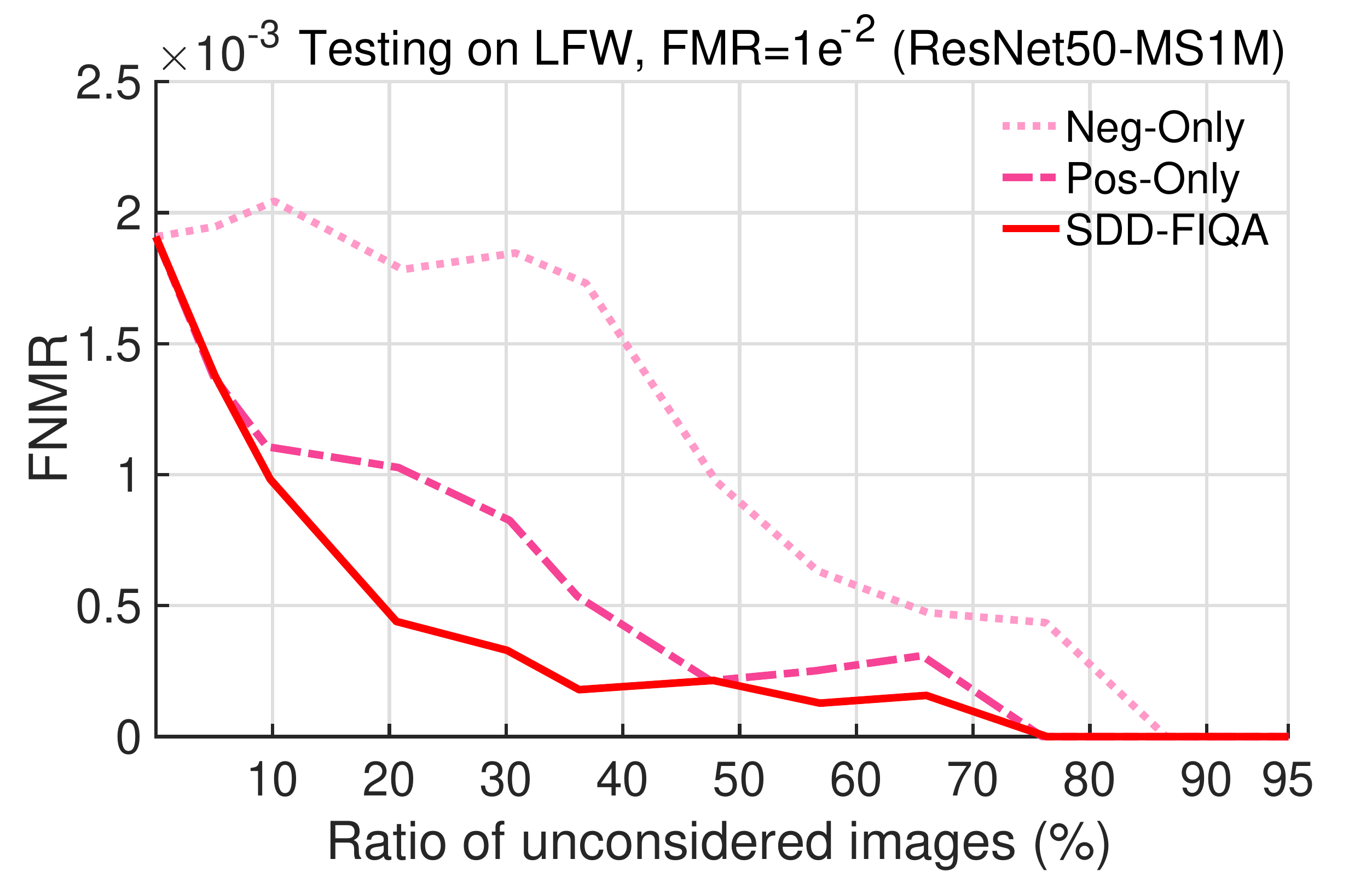}
\caption{EVRC comparison of using Pos-Only, Neg-Only, and the proposed  SDD-FIQA on the LFW dataset.}
\label{ablation}
\end{figure}

\noindent{\textbf{Similarity distribution information comparison}}. To confirm the efficacy of utilizing both positive pairs and negative pairs, we report the EVRC results obtained by three different generation schemes of quality pseudo-labels: 1) only using positive pairs similarity distribution information (Pos-Only) by setting the $S^N_{x_i}$ to 1 in Eq.~\eqref{eq3.4}, 2) only using negative pairs similarity distribution information (Neg-Only) by setting the $S^P_{x_i}$ to -1 in Eq.~\eqref{eq3.4}, and 3) using the similarity distribution distance between the positive pairs and negative pairs (SDD-FIQA). By observing the curves in Fig. \ref{ablation}, we can see that utilizing both intra-class and inter-class information can achieve the best result, and only using intra-class information achieves a better result than only using inter-class information. It is worth mentioning that Neg-Only can also reduce FNMR, which proves the fact that the inter-class information is also effective to FIQA.

\begin{table}[]
\setlength{\abovecaptionskip}{0.05cm}
\setlength{\belowcaptionskip}{-0.1cm}
\begin{center}
\caption{AOC results on cross recognition model setting. The deployed recognition is ResNet101-CASIA.}\label{ACC_Res101_CASIA}
\resizebox{236pt}{135pt}{
\begin{tabular}{|C{0.7cm}|c|c|c|c|c|}
\Xhline{0.6pt}
$ $	 &\bf{LFW}	&\bf{FMR}=$\bm{1e^{-2}}$ 	&\bf{FMR=}$\bm{1e^{-3}}$		&\bf{FMR=}$\bm{1e^{-4}}$ 	&\bf{Avg}\\
\Xhline{0.6pt}
\multirow{3}*{\rotatebox{90}{\makecell{\textbf{\small{Analytics}}\\ \textbf{\small{Based}}}}}
& BRISQUE \cite{BRISQUE}				& 0.0689   &0.0976   &0.1870   &0.1178\\
& BLIINDS-II \cite{BLIINDS-II}		&-0.0065   &0.0557   &0.2277   &0.0923 \\
& PQR \cite{PQR}						& 0.1771   &0.0213   &0.0849   &0.0944 \\
\hline
\multirow{5}*{\rotatebox{90}{\makecell{\textbf{\small{Learning}}\\ \textbf{\small{Based}}}}}
& FaceQnet-V0 \cite{faceqnetv02019} 	& 0.5454   &0.4635   &0.4649   &0.4913 \\ 
& FaceQnet-V1 \cite{faceqnetv12020} 	& 0.5938   &0.5174   &0.4842   &0.5318 \\
& PFE \cite{PFE2019}					& 0.6381   &0.6500   &0.6090   &0.6324 \\ 
& SER-FIQ \cite{SER-FIQ2020} 		& 0.6212   &0.5413   &0.4962   &0.5529 \\ 
& PCNet \cite{PCNet2020} 		&0.6629  &0.6398   &0.6079   &0.6369 \\
\Xhline{0.6pt}
& \bf{SDD-FIQA (Our)} 						& \bf{0.7395}   &\bf{0.7001}   &\bf{0.6296}   &\bf{0.6898} \\ 
\hline
\hline
\hline
$ $	 &\bf{Adience}	&\bf{FMR}=$\bm{1e^{-2}}$ 	&\bf{FMR=}$\bm{1e^{-3}}$		&\bf{FMR=}$\bm{1e^{-4}}$ 	&\bf{Avg}\\
\Xhline{0.6pt}
\multirow{3}*{\rotatebox{90}{\makecell{\textbf{\small{Analytics}}\\ \textbf{\small{Based}}}}}
& BRISQUE \cite{BRISQUE}				&0.1551   &0.1398   &0.1489   &0.1479 \\
& BLIINDS-II \cite{BLIINDS-II}		&0.1163   &0.1037   &0.1337   &0.1179 \\
& PQR \cite{PQR}						&0.1559   &0.1327   &0.1140   &0.1342 \\
\hline
\multirow{5}*{\rotatebox{90}{\makecell{\textbf{\small{Learning}}\\ \textbf{\small{Based}}}}}
& FaceQnet-V0 \cite{faceqnetv02019} 	&0.4244   &0.3271   &0.2840   &0.3452 \\ 
& FaceQnet-V1 \cite{faceqnetv12020} 	&0.4283   &0.3136   &0.2524   &0.3314 \\
& PFE \cite{PFE2019}					&0.5730   &0.4392   &0.3154   &0.4425 \\
& SER-FIQ \cite{SER-FIQ2020}  		&0.4529   &0.3327   &0.2826   &0.3561 \\
& PCNet \cite{PCNet2020} 		&0.5178  &0.3935   &0.2962   &0.4025 \\
\hline
& \bf{SDD-FIQA (Our)} 						&\bf{0.5790}   &\bf{0.4535}   &\bf{0.3443}   &\bf{0.4589} \\ 
\hline
\hline
\hline
$ $	 &\bf{IJB-C}	&\bf{FMR}=$\bm{1e^{-2}}$ 	&\bf{FMR=}$\bm{1e^{-3}}$		&\bf{FMR=}$\bm{1e^{-4}}$ 	&\bf{Avg}\\
\Xhline{0.6pt}
\multirow{3}*{\rotatebox{90}{\makecell{\textbf{\small{Analytics}}\\ \textbf{\small{Based}}}}}
& BRISQUE \cite{BRISQUE}				&0.4014   &0.3680   &0.3769   &0.3821 \\
& BLIINDS-II \cite{BLIINDS-II}		&0.3202   &0.2276  &0.2920  &0.2799\\
& PQR \cite{PQR}						&0.4747   &0.4047   &0.3304   &0.4033 \\
\hline
\multirow{5}*{\rotatebox{90}{\makecell{\textbf{\small{Learning}}\\ \textbf{\small{Based}}}}}
& FaceQnet-V0 \cite{faceqnetv02019} 	&0.5469   &0.4212   &0.3803   &0.4495 \\ 
& FaceQnet-V1 \cite{faceqnetv12020} 	&0.5411  &0.4742   &0.3983   &0.4712 \\
& PFE \cite{PFE2019}					&0.6674   &0.5512   &\bf{0.4848}   &0.5678 \\
& SER-FIQ \cite{SER-FIQ2020}  		&0.5453   &0.4276    &0.4190   &0.4640 \\ 
& PCNet \cite{PCNet2020} 		&0.6537   &0.5379   &0.4619   &0.5512 \\ 
\Xhline{0.6pt}
& \bf{SDD-FIQA (Our)} 						&\bf{0.6768}   &\bf{0.5628}   &0.4832   &\bf{0.5743} \\ 
\Xhline{0.6pt}

\end{tabular}}
\end{center}
\end{table}

\section{Conclusion}

This paper proposes a novel FIQA approach called SDD-FIQA, which considers recognition Similarity Distribution Distance (SDD) as face image quality annotation. The novelties of our algorithm are three-fold: First, we are the first to consider the FIQA as a recognizability estimation problem. Second, we propose a new framework to map the intra-class and inter-class similarity to quality pseudo-labels via the Wasserstein metric, which is closely related to the recognition performance. Third, an efficient implementation of SDD is developed to speed up the label generation and reduce the time complexity from $\mathcal{O}(n^2)$ to $\mathcal{O}(n)$. Compared with existing FIQA methods, the proposed SDD-FIQA shows better accuracy and generalization.

{\small
\bibliographystyle{ieee_fullname}
\bibliography{egbib}
}

\end{document}